\documentclass[10pt,journal]{IEEEtran}
\usepackage{cite}
\usepackage{amsmath,amssymb,amsfonts}
\usepackage{algorithmic}
\usepackage{graphicx}
\usepackage{multirow}
\usepackage{textcomp}
\usepackage{xcolor}
\usepackage{graphicx}
\usepackage{subfigure}
\usepackage{CJKutf8}
\usepackage{algorithm}
\usepackage{algorithmic}
\usepackage{etoolbox}

\newtheorem{assumption}{Assumption}
\newenvironment{sloppypar*}
{\sloppy\ignorespaces}
{\par}
\def\BibTeX{{\rm B\kern-.05em{\sc i\kern-.025em b}\kern-.08em
    T\kern-.1667em\lower.7ex\hbox{E}\kern-.125emX}}
\begin{document}\sloppy

\title{Robust Semi-supervised Federated Learning for Images Automatic Recognition in Internet of Drones}

\author{Zhe Zhang\textsuperscript{$\dagger$},
        Shiyao Ma\textsuperscript{$\dagger$},
        Zhaohui Yang,
        Zehui Xiong,
        Jiawen Kang,
        Yi Wu,
        Kejia Zhang,
        Dusit Niyato,~\IEEEmembership{Fellow,~IEEE}
\thanks{Z. Zhang and Y. Wu are the School of Data Science and Technology, Heilongjiang University, China (e-mail: zhangzhe97cs@163.com, 1995050@hlju.edu.cn). S. Ma is with the College of Information and Communication Engineering, Dalian Minzu University, China (e-mail: shiyaoma.cs@gmail.com). Z. Yang is with the Department of Electronic and Electrical Engineering, University College London, WC1E 6BT London, UK (e-mail: zhaohui.yang@ucl.ac.uk). Z. Xiong is with the Information Systems Technology and Design Pillar, Singapore University of Technology and Design, Singapore (e-mail: zehui\_xiong@sutd.edu.sg). J. Kang is with the Automation School, Guangdong University of Technology, China (e-mail: kjwx886@163.com). K. Zhang is with the School of Mathematical Science, Heilongjiang University (e-mail: zhangkejia@hlju.edu.cn). D. Niyato is with School of Computer Science and Engineering, Nanyang Technological University, Singapore (e-mail: dniyato@ntu.edu.sg).}
\thanks{\textsuperscript{$\dagger$}Equal contributions and Y. Wu is the corresponding author.}}

\markboth{IEEE Internet of Things Journal}%
{Shell \MakeLowercase{\textit{et al.}}: Bare Demo of IEEEtran.cls for IEEE Journals}

\maketitle

\begin{abstract}
Air access networks have been recognized as a significant driver of various Internet of Things (IoT) services and applications.
In particular, the aerial computing network infrastructure centered on the Internet of Drones has set off a new revolution in automatic image recognition.
This emerging technology relies on sharing ground truth labeled data between Unmanned Aerial Vehicle (UAV) swarms to train a high-quality automatic image recognition model. However, such an approach will bring data privacy and data availability challenges. To address these issues, we first present a Semi-supervised Federated Learning (SSFL) framework for privacy-preserving UAV image recognition.
Specifically, we propose model parameters mixing strategy to improve the naive combination of FL and semi-supervised learning methods under two realistic scenarios (\emph{labels-at-client} and \emph{labels-at-server}), which is referred to as Federated Mixing ($\mathrm{FedMix}$).
Furthermore, there are significant differences in the number, features, and distribution of local data collected by UAVs using different camera modules in different environments, i.e., statistical heterogeneity.
To alleviate the statistical heterogeneity problem, we propose an aggregation rule based on the frequency of the client's participation in training, namely the $\mathrm{FedFreq}$ aggregation rule, which can adjust the weight of the corresponding local model according to its frequency.
Numerical results demonstrate that the performance of our proposed method is significantly better than those of the current baseline and is robust to different non-IID levels of client data.

\end{abstract}

\begin{IEEEkeywords}
Federated learning, semi-supervised learning, non-IID, unmanned
aerial vehicle (UAV), aerial computing.
\end{IEEEkeywords}

\section{Introduction}
Advances in aerial computing have made many emerging applications possible, such as, aerial photography and infrastructure inspection, which have put an increasing impact on human daily life \cite{liu2020secure,9481289,9094657,lim2021uav}.
Those applications are deployed on various mobile terminals and IoT devices \cite{liu2020privacy}, and the most widely of which is automatic image recognition based on the Internet of Drones.
\textcolor{black}{Compared with traditional ground photography, low-altitude unmanned aerial vehicles (UAVs) can usually flexibly capture clear and complete image data to achieve tasks such as vehicle classification on the road and wildlife monitoring in the natural environment.}
Therefore, aerial computing empowered UAV has spawned many new intelligent applications and has become a promising new computing paradigm.

However, it generally requires a large amount of data with ground-truth labels to train a high-quality recognition model \cite{9205981}. 
%
%
Meanwhile, in most cases, gathering together an adequately labeled dataset is a time-consuming, expensive, and complicated endeavor \cite{liu2020rc}.
%
%
\textcolor{black}{On the other hand, the images captured by UAVs usually contain private information such as the user's behavior trajectory and location, which makes it impossible to share raw data with other companies.} \textcolor{black}{For example, GDPR \cite{voigt2017eu} stipulated that all organizations cannot share users' private data without permission because it reveal their privacy.}
Therefore, designing an AI framework model by multiple participants to meet the regulatory privacy requirements is a promising solution to solve the above problems.


Federated Learning (FL) \cite{mcmahan2017communication,lim2021decentralized} is essentially a distributed privacy-preserving machine learning framework, which allows participants to hold data locally instead of sharing data to train a shared global model collaboratively. 
In FL, clients train a model from their local data samples, and the server only aggregates data holders' local model updates for data privacy preservation \cite{9277438,liu2021towards}.
Motivated by this fact, our insight is to apply the FL framework to the Internet of Drones to solve the problem of data privacy.
However, such a method still faces the following challenges:

\begin{itemize}
\item\textbf{\textit{Lack of labeled data:}} The current mainstream work is based on an unrealistic assumption: \emph{the training data of local clients have ground-truth} \cite{liu2020rc,9094657}. However, the UAV's local dataset has only a few or no labels in real scenarios. This phenomenon of missing labels is usually caused by high annotating costs or a lack of expert knowledge in related fields.

\item\textbf{\textit{Statistical heterogeneity:}} Different types of UAVs use different camera modules to collect local data in different environments (e.g., sunny days and thunderstorms) \cite{9184079}. In particular, the local dataset of each client may have different distributions and volumes for each category \cite{9514368}. As a result, there are too many differences in data distribution, features, and the number of labels between clients, which is not conducive to the convergence of the global model.
\end{itemize}

To respond to the first challenge, researchers generally use Semi-supervised Learning (SSL) methods (such as consistency regularization \cite{samuli2017temporal,park2018adversarial} and pseudo-label \cite{lee2013pseudo}) to train AI models under setting lacking labeled data.
Inspired by SSL, recent works \cite{liu2020rc,long2020fedsemi,jin2020towards,jeong2021federated,9134408} studied how to design a Semi-supervised Federated Learning (SSFL) framework, which can effectively integrate semi-supervised learning into the FL framework.
For example, Liu \textit{et al.} in \cite{liu2020rc} utilized pseudo-labels whose predicted value is higher than the confident threshold to train SSFL models, where the performance of the model determines the predicted value of the sample. However, in this method, the model performs poorly in the early stages of training, resulting in the generated pseudo labels relatively low quality. To this end, we introduce the hyperparameter items that dynamically change with \textcolor{black}{iteration rounds} to adaptively adjust the influence of the pseudo-label method on the local model.
Meanwhile, to save computing costs, we use active learning strategies to filter out a high-quality subset of pseudo-label samples instead of all samples to train the local model.

Furthermore, previous SSFL work, such as $\mathrm{FedMatch}$ \cite{jeong2021federated} only focused on how to decompose the parameters of labeled and unlabeled data for disjoint learning. In this way, the learned global model will be biased towards labeled data (supervised model) or unlabeled data (unsupervised model) instead of overall data (global model). 
Thus, we further propose model parameters mixing strategy for disjoint learning of supervised model (learned on labeled data), unsupervised model (learned on unlabeled data), and global model.

To respond to the second challenge, researchers have proposed many robust learning techniques in FL. For example, $\mathrm{FedProx}$ \cite{li2020federatedprox} introduced an additional $\ell_2$ regularization term in the local objective function so that the average model slowly approaches the global optimum. $\mathrm{FedBN}$ \cite{li2020fedbn} utilized local batch normalization to alleviate the feature shift before average aggregating local models.
However, methods such as these add additional computational and communication overhead to the server or client \cite{liu2021resource}. Considering the limited storage, computing, and communication capabilities of the node devices (i.e., the drones on Internet of Drones systems) in FL, we must explore an effective method to alleviate the non-IID problem. To this insight, we propose a robust aggregation rule, which dynamically adjusts the corresponding local model's weight by recording the client's training frequency.
Moreover, we introduced the Dirichlet distribution function to simulate the different non-IID levels of client data. Therefore, the main contributions of this paper are summarized as follows:
\begin{itemize}
    \item
    To address data privacy leakage in the aerial computing framework, we apply FL framework to the aerial computing framework to enhance data privacy protection capabilities. In particular, we focus on computer vision tasks on the Internet of Drones and propose an FL algorithm that enables UAV swarms \textcolor{black}{different companies} to achieve high-precision image recognition without sharing the raw data.
    \item
    To solve data availability in FL, we propose a robust semi-supervised FL system, which performs model parameters mixing strategy of disjointed learning for the supervised, unsupervised, and global models. \textcolor{black}{Meanwhile, we designed a dynamic hyperparameter item and two active learning strategies in the SSFL system to achieve high-quality performance.}
    \item
    To handle statistical heterogeneity in FL, \textcolor{black}{we propose a new model aggregation rule without increasing the additional computational overhead of UAVs,} which dynamically adjusts the corresponding local model's weight by recording the client's training frequency to alleviate the non-IID problem.
    \item
    We experimentally evaluate the feasibility of the SSFL system for standard image classification tasks with CIFAR-10 and Fashion-MNIST datasets. The simulation evaluation results show that the system performance we designed is better than the popular baselines.
\end{itemize}

\section{related work}\label{sec-2}
\subsection{Privacy-protected UAV Image Automatic Recognition}
In recent years, the Internet of Drones has been widely used in many fields such as pedestrian tracking, atmospheric monitoring, and geological prospecting due to its excellent mobility and flexibility \cite{9277438}. In particular, the applications mentioned above all rely on the automatic recognition of the collected UAV images by the UAV swarm. 
\textcolor{black}{However, images captured by drone swarms are usually related to the privacy of users, requiring data to be stored locally and cannot be shared with each other.
Researchers apply the FL framework to the Internet of Drones to solve the problem of data privacy.}
For instance, Liu \textit{et al.} in \cite{9184079} proposed a long-term, short-term memory UAV recognition model based on graph convolutional neural network and FL to achieve accurate and real-time inference of air quality index. Lim \textit{et al.} in \cite{9354588} proposed an FL-based sensing and collaborative learning approach for UAV-enabled internet of vehicles (IoVs), where UAVs collect data and train Machine Learning (ML) models for IoVs. While these methods improve UAV image automatic recognition efficiency and accuracy, they still face two serious challenges: data availability and data privacy.

\subsection{Semi-supervised Federated Learning}
Semi-supervised Federated Learning attempts to use semi-supervised learning techniques \cite{zhu2009introduction,chapelle2009semi,kingma2014semi,zhai2019s4l,mallapragada2008semiboost} to further improve the performance of FL models trained in realistic scenarios with insufficient labeled data.
For the research of SSFL, previous work mainly focused on the method of naive integration of SSL into FL. In their pioneering study, the SSFL was divided into two scenarios: standard and disjoint. While the former scenario is characterized by the labeled data being on the client-side, the latter is a more challenging scenario in which the labeled data is only available on the server-side.
Then the semi-supervised learning method (i.e., consistent regularization and pseudo-labels) is used to train the FL model on the labeled and unlabeled datasets, respectively. These works provide an essential solution for the SSFL image classification task. However, it is still necessary to explore the relationship between data and model and maximize the use of unlabeled data to improve model performance.

Therefore, some researchers designed independent learning of supervised and unsupervised models and improve semi-supervised learning methods to achieve high-performance FL models. For example, Jeong \textit{et al.} in \cite{jeong2021federated} proposed a new inter-client consistency loss and model parameter decomposition strategy for non-joint learning of labeled and unlabeled data.
Long \textit{et al.} in \cite{long2020fedsemi} proposed a semi-supervised federated learning system: \textit{FedSemi}, which unifies the consistency-based semi-supervised learning model \cite{lee2013pseudo}, dual model \cite{samuli2017temporal}, and average teacher model \cite{tarvainen2017mean} to achieve SSFL. The DS-FL system proposed in \cite{9392310} aims to solve the communication overhead problem in semi-supervised federated learning.

\subsection{Robust Federated Learning}
Since the local dataset distribution of each client is different from the global distribution, the local loss function of the client is inconsistent with the global optimal \cite{li2021federated}, resulting in poor model performance. Especially when the local client model parameters update are large, the difference will be more obvious. Therefore, in order to alleviate the non-IID problem (i.e., the local dataset distribution of each client is inconsistent) \cite{li2019convergence,zhao2018federated,sattler2019robust,briggs2020federated,li2020fedbn,wang2020optimizing,chen2020asynchronous}, researchers have made many efforts in the FL field.
Some studies design a robust federated learning algorithm to solve the non-IID problem. For example, \textit{FedProx} \cite{li2020federatedprox} limits the distance between the local model and the global model by introducing an additional $\ell_2$ regularization term in the local target function to limiting the size of the local model update. However, this method has a disadvantage in that each client needs to individually adjust the local regularization term to obtain good model performance. \textit{FedNova} \cite{wang2020tackling} improved FedAvg in the aggregation phase, which normalized and scaled the model update according to the local training batch of the client. 
Although previous studies have alleviated the problem of non-IID to some extent, none of these researchers took into account the problems of data heterogeneity and insufficient data simultaneously.

\textcolor{black}{Inspired by previous work, this paper considers insufficient labeled data and statistical heterogeneity faced by drone groups with privacy protection in performing aerial image recognition tasks.}
To this end, we propose a robust semi-supervised federated learning system to solve the above challenges, which can achieve high recognition accuracy under the setting of lack of labeled data and non-IID distribution.

\section{Preliminaries}\label{sec-3}
\subsection{Federated Learning}
Federated Learning (FL) \cite{mcmahan2017communication} solves the problem of data island on the premise of privacy protection.
In particular, the FL is a distributed machine learning framework, which requires clients to hold data locally, where these clients coordinate to train a shared global model $\omega^\ast$.
In FL, there is a server $\mathcal{S}$ and $K$ clients, each of which holds an IID or non-IID datasets $\mathcal{D}_k$.
Specifically, for a training sample $x$ on the client side, let $\ell(\omega; x)$ be the loss function at the client, where $\omega  \in {\mathbb{R}^d}$ denotes the model’s trainable parameters.
Therefore, we let $\mathcal{L}(\omega) = \mathbb{E}_{x \sim \mathcal{D}}[\ell(\omega ; x)]$ be the loss function at the server. Thus, FL needs to optimize the following objective function at the server:
\begin{equation}
	{\min _\omega }\mathcal{L}(\omega ),{\text{ where }}\mathcal{L}(\omega ) = \sum\limits_{k = 1}^K {{p_k}} {\mathcal{L}_k}(\omega ),
\end{equation}
where ${p_k\ge 0}$, $\sum\limits_k {{p_k}}  = 1$ indicates the relative influence of $k$-th client on the global model.
Typically, the FL system uses the federated average (FedAvg \cite{mcmahan2017communication}) algorithm, which includes three stages: initialization, local training, and server aggregation.
The detailed training steps of FedAvg are summarized as follows:
\begin{itemize}
    \item \textbf{\textit{Step 1, Initialization:}} In $t$-th round of training, the server randomly selects a subset of clients from all participating clients, i.e., $\mathcal{K}_t \subseteq \mathcal{K}$. After that, the server sends the initialized global model $\omega_t$ to the selected clients.
    \item \textbf{\textit{Step 2, Local training:}} The client uses the local optimizer (e.g., SGD, Adam) on the local dataset $\mathcal{D}_k$ to train the received initialization model. For the $k$-th client, the following objective function should be minimized:
    \begin{equation}
	{\mathcal{L}_k}(\omega_{t}^{k}) = \dfrac{1}{|D_k|} \sum\nolimits_{{x_i},{y_i} \in {{\cal D}_k}} \ell_{i} (y_{i}, f_{\omega_{k}}(x_{i})),
    \end{equation}
    where $f_{\omega_{k}}(x_{i})$ indicates the model output of  $x_i$. Then, each client uploads the local model $\omega_{t}^{k}$ to the server.
    \item \textbf{\textit{Step 3, Aggregation:}} The server collects and uses the federated average algorithm to aggregate the models uploaded by these clients to obtain a new global model, i.e., 
    \begin{equation}
    \omega_{t+1} = \dfrac{1}{|\mathcal{K}_t|} \sum\nolimits_{k \in {\mathcal{K}_t}} \omega_{t}^{k}.
    \end{equation} 
Next, the server broadcasts the updated global model $\omega_{t+1}$ to all selected clients.
\end{itemize}

Note that FL repeats the above steps until the global model converges.

\begin{figure}[!t]
	\centering
	\includegraphics[width=1\linewidth]{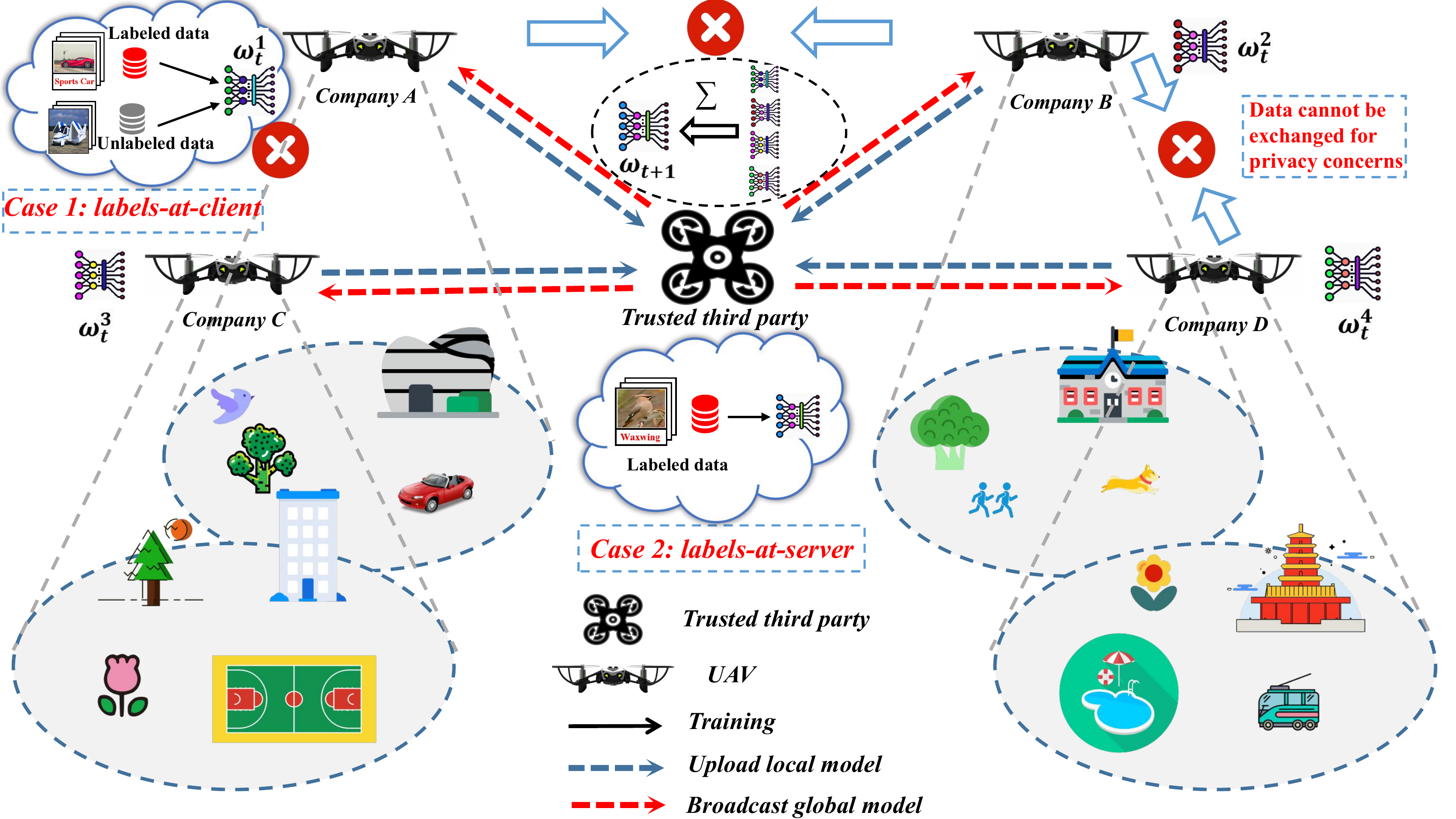}
	\caption{\textcolor{black}{Overview of semi-supervised federated learning system under UAV aerial image recognition.}}\label{system}
\end{figure}

\subsection{Semi-supervised Learning}
In the real world (e.g., financial and medical fields), unlabeled data is easy to gain, while labeled data is often difficult to obtain. 
Meanwhile, annotating data requires a lot of human resources and material resources. \textcolor{black}{To this end, the researchers proposed a machine learning paradigm, namely semi-supervised learning \cite{sajjadi2016regularization,xie2020unsupervised}, which can learn a model close to the performance of a fully-supervised model on a dataset containing unlabeled samples.
Thus, semi-supervised learning has become a panacea to solve the problem of data availability.
Specifically, we are given a dataset $D = \mathcal{X} \cup \mathcal{U}$ and an initialized model $\theta$, where $\mathcal{X} = \{(x_j, y_j)\}_{j=1}^{n}$, $\mathcal{U} =  \{u_i\}_{i=1}^{m}$. Here, in general, $m\gg n$.
For the labeled dataset $\mathcal{X}$, we use the supervised learning method to train the model $\theta$.
For the unlabeled dataset $\mathcal{U}$, we use the supervised learning method to continue training model $\theta$.
Next, we introduce in detail a fundamental assumption and two common methods of semi-supervised learning.}

\begin{assumption}
\textit{In machine learning, there is a basic assumption that if the features of two unlabeled samples $u_1$ and $u_2$ are similar, the corresponding model prediction results $y_1$ and $y_2$ are the same \cite{yang2021survey}, i.e., $f({u_1}) = f({u_2})$, where $f( \cdot )$ is the prediction function.}
\end{assumption}

\textbf{Consistency Regularization}: The main idea of this method is that the model prediction results should be the same whether noise is added or not on an unlabeled training sample \cite{sajjadi2016regularization,samuli2017temporal}. 
We generally use data augmentation (such as image flipping and shifting) methods to add noise to increase the diversity of the dataset. 
Specifically, for an unlabeled sample $u_i$ and its perturbation form $\hat{u_i}$, our goal is to minimize the distance $d (f_\theta (u_i),f_\theta (\hat{u_i}))$, where $f_\theta (u_i)$ is the output of sample $u_i$ on model $\theta$.
The common distance measurement method is Kullback-Leiber (KL) divergence.
Thus, the loss of the consistency regularization method is defined as follows:
\begin{equation}
d (f_\theta; u)_{KL} = \dfrac{1}{m} \sum\limits_{i = 1}^m f_\theta (u_i) log\dfrac{f_\theta (u_i)}{f_\theta (\hat{u_i})},
\end{equation}
where $m$ is the total number of unlabeled samples and $f_\theta (u_i)$ indicates the model output of unlabeled sample $u_i$.


\textbf{Pseudo-label}: 
\textcolor{black}{The pseudo-label method \cite{lee2013pseudo} is to utilize the model $\theta$ to set pseudo-labels on unlabeled data.
In this method, pseudo labels refer to picking up the class which has the maximum predicted probability, are used as if they were true labels.
Specifically, for an unlabeled sample $u_i$, we assume its pseudo label is $\widehat y_i$, where $\widehat{y_i}=(\widehat{y_{i}^{1}}, \cdots, \widehat{y_{i}^{c}})$, $c$ is the total number of sample classes.
The pseudo label is calculated as follows:
\begin{equation}
    \widehat{y_{i}^{c}} = \begin{cases}1&{if}\ c = \mathop{\arg\max}_{c}\ (f_\theta (u_i)_c)\\0&{otherwise}\end{cases},
\end{equation}
where $f_\theta (u_i)_c$ denotes the predicted probability value of belong to the $c$-th class of $f_{\theta}(u_i)$.
After that, for the pseudo-labeled dataset $\mathcal{\widehat{U}} =  \{u_i, \widehat y_i\}_{i=1}^{m}$ , we use the supervised learning method to train model $\theta$.
Moreover, the summaries of symbols are presented in Table \ref{tab-n}.}

\begin{table}[!t]
\scriptsize
	\centering
	\caption{\textcolor{black}{Summaries of applied symbols.}}
	\begin{tabular}{|c|c|}\hline
		Symbols & Description\\\hline
		$\mathcal{X}$ & The labeled dataset\\
		$\mathcal{U}$ & The unlabeled dataset\\
		$\mathcal{D}_s$ & The server local dataset\\
		$\mathcal{D}_k$ & The $k$-th client local dataset\\
		$\mathcal{K}_t$ & The set of clients selected in round $t$\\
		$x_i$ & The $i$-th labeled sample\\
		$u_i$ & The $i$-th unlabeled sample\\
		$y_i$ & The label of the $i$-th sample\\
		$\widehat{y_i}$ & The pseudo label of the $i$-th sample\\
		$f_\theta ( \cdot )$ & The prediction function on model $\theta$\\
	    $\pi (\cdot)$ & The form of perturbation\\
		$\omega$ & The global model\\
		$\sigma$ & The supervised model\\
		$\psi$ & The unsupervised model\\
		$\alpha$ & The weight of unsupervised model\\
		$\beta$ & The weight of supervised model\\
		$\gamma$ & The weight of global model\\
		$F$ & The client participation rate ($0<F<1$)\\
		$K$ & The total number of clients\\
		$B$ & The local min-batch size\\
		$E$ & The number of local epochs\\
		$t$ & The round of iteration\\
        $c$ & The total number of classes\\
		$n$ & The number of pseudo labels\\
		$q_{t+1}^{k}$ & The number of times that the $k$-th client\\
		{} & has been trained up to the $t+1$-th round\\
		$w_{t}^{k}$ & The weight of the $k$-th client in round $t$\\
		$\mu$ & The parameters of Dirichlet distribution function\\\hline
	\end{tabular}
	\label{tab-n}
\end{table}

\section{Problem Definition}\label{sec-4}
Most of the current works focus on training models based on labeled data. However, due to high annotating costs, the UAV's local dataset usually has only a few labeled data or even no labeled data. Semi-supervised learning methods can use unlabeled data to alleviate the dependence of the model on labeled data. Therefore, we apply the semi-supervised learning methods to the UAV framework. Due to the data privacy issues, we introduce these methods in an FL-based UAV framework called the Semi-supervised Federated Learning-based (SSFL) UAV framework. \textcolor{black}{As shown in Fig. \ref{system}, we regard the leading UAV of a trusted third party as the server, and other UAV swarms as the clients.} Furthermore, there are two essential scenarios of SSFL based on the location of the labeled data. The first scenario considers a conventional case where UAVs have both labeled and unlabeled data (i.e., \emph{labels-at-client}), and the second scenario assumes that the labeled data is only available at the server (i.e., \emph{labels-at-server}), which is more challenging.
Next, we define the problem studied in this paper as follows:

\noindent \textbf{Labels-at-Client Scenario}:
\textcolor{black}{Suppose that drones of different companies perform vehicle classification tasks at high altitudes.
However, these companies may not want to spend too much time and effort annotating the data captured by drones, which will leave most of the data unlabeled.\textbf{(See Case 1 in Fig. \ref{system})}.
Thus,} in this scenario, we assume that there are $K$ clients and a server $\mathcal{S}$, where the server does not hold any data while each client holds a local dataset $\mathcal{D}_k$, where $\mathcal{D}_k=\mathcal{D}_{k}^{s} + \mathcal{D}_{k}^{u}$, where the labeled dataset and unlabeled dataset of the $k$-th client are $\mathcal{D}_{k}^{s}=\{(x_j, y_j)\}_{j=1} ^{n}$ and $\mathcal{D}_{k}^{u}=\{u_i\}_{i=1} ^{m}$, respectively.
In the training process, the client should consider the loss of classification of labeled data and the loss of consistency of unlabeled data to train the local model. Therefore, the loss function $\mathcal{L}^k$ of the $k$-th client is defined as:
\textcolor{black}{\begin{equation}
\begin{aligned}
{\cal L}^k &= \frac{1}{n}\sum\nolimits_{{x_j} \in {{\cal D}_{k} ^ {s}}} C E({y_j},{f_{{\theta _k}}}({x_j}))\\
& + \dfrac{1}{m} \sum\nolimits_{{u_i} \in {{\cal D}_{k}^{u}}} {f_{{\theta _k}}}({u_i}) log\dfrac{{f_{{\theta _k}}}({u_i})}{{f_{{\theta _k}}}(\pi ({u_i}))},
\end{aligned}
\end{equation}}
where $n$ and $m$ are the numbers of labeled samples and unlabeled samples, respectively, $\pi(\cdot)$ is the data augmentation function (e.g., flip and shift of the unlabeled samples), $f_{\theta_k} (x_i)$ indicates the output of labeled sample $x_i$ on model $\theta_k$ of the $k$-th client. For the labels-at-client scenario, the client trains on labeled and unlabeled data. Like FL, the server only aggregates the updates obtained from the client and resends the aggregated model parameters to the client.

\noindent \textbf{Labels-at-Server Scenario}: \textcolor{black}{
Assume that UAVs of different companies perform automatic image classification tasks of wild birds in a natural environment.
In this case, the company may not have enough expertise to properly annotate images of wild birds captured by drones. 
Therefore, these images are unlabeled.
However, third-party trusted agencies (such as the government) can store some labeled wild bird images \textbf{(See Case 2 in Fig. \ref{system})}.
Thus, in this scenario, the server holds a labeled dataset $\mathcal{D}_s=\{(x_i, y_i)\}_{i=1} ^{n}$ and each client holds a local unlabeled dataset $\mathcal{D}_{k}^{u}=\{u_i\}_{i=1} ^{m}$. Moreover, unlike the label-at-client scenario, the client only needs to consider the unsupervised loss.} Therefore, let $\mathcal{L}_u^k$ be defined as the loss function of the $k$-th client:
\textcolor{black}{\begin{equation}
\begin{aligned}
{\cal L}_u^k &= \frac{1}{m}\sum\nolimits_{{u_i} \in {{\cal D}_k}} C E(\widehat {{y_i}},{f_{{\theta _k}}}({u_i}))\\
& + \frac{1}{m}\sum\nolimits_{{u_i} \in {{\cal D}_k}} {f_{{\theta _k}}}({u_i}) log\dfrac{{f_{{\theta _k}}}({u_i})}{{f_{{\theta _k}}}(\pi ({u_i}))},
\end{aligned}
\end{equation}}
where $m$ is the number of unlabeled samples, $\pi(\cdot)$ is the data augmentation function, $\hat{y_i}$ is the pseudo label of unlabeled sample $u_i$, and $f_{\theta_k} (u_i)$ indicates the output of unlabeled sample $u_i$ on model $\theta_k$ of the $k$-th client.
Then the server uses aggregation algorithm to aggregate the client's model parameters to obtain the global model $\theta$, i.e., $\theta=\dfrac{1}{K}\sum\limits_{k = 1}^K \theta
_{k}$.
Furthermore, let $\mathcal{L}_s$ be defined as the loss function at the server side:
\begin{equation}
{{\cal L}_s} = \frac{1}{n}\sum\nolimits_{{x_i},{y_i} \in {{\cal D}_s}} C E({y_i},{f_\theta }({x_i})),
\end{equation}
where $n$ is the number of labeled samples, and $f_\theta (x_i)$ indicates the output of labeled sample $x_i$ on model $\theta$.
Therefore, the objective function of this scenario in SSFL system is to minimize the following loss function: 
\begin{equation}\label{eq-5}
\min {\cal L},\text{where} \quad {\cal L} \buildrel\textstyle.\over= \sum\limits_{k = 1}^K {{\cal L}_u^k}  + {{\cal L}_s}.
\end{equation}

Note that the whole learning process is similar to the traditional FL system, except that the server not only aggregates the client model parameters but also trains the model with labeled data.

\section{Algorithm and System Design}\label{sec-5}
\subsection{Semi-supervised Federated Learning System Design}
Similar to the traditional FL system, in our SSFL-UAV framework, the leading UAV and other UAV swarms are cooperative to train a high-performance global model $\omega^*$. 
\textcolor{black}{The previous work \cite{liu2020rc,long2020fedsemi,jeong2021federated} used the FedAvg algorithm to directly aggregate the models uploaded by the client.
By this way, the aggregated average model may be far from the global optima especially when the local updates are large (e.g., a large number of local epochs) \cite{li2020federatedprox}.
This is because they ignore the implicit contribution between iterations of the global model.}
Moreover, in the standard SSL method, the learning of labeled data and unlabeled data is completed on a model, which may cause the model to forget the knowledge learned from the labeled data \cite{serra2018overcoming}.
\textcolor{black}{The reason for this phenomenon is that the amount of unlabeled data is far greater than labeled data.}
Inspired by the above facts, based on UAV aerial image recognition, we propose an SSFL algorithm called $\mathrm{FedMix}$ that focuses on the implicit contributions between iterations of the global model in a fine-grained manner. 
To realize the separated learning of labeled and unlabeled data, we define the supervised model trained on the labeled dataset as $\sigma$, the unsupervised model trained on the unlabeled dataset as $\psi$, and the aggregated global model as $\omega$. Specifically, we design a strategy that assigns three weights $\alpha,\beta,$ and $\gamma$ to the unsupervised model $\psi$, supervised model $\sigma$, and the previous round of global model, respectively. The designed algorithm can capture the implicit relationship between each iteration of the global model in a fine-grained manner. Next, we describe in detail the core components of the $\mathrm{FedMix}$ algorithm.

\begin{figure}[!t]
	\centering
	\includegraphics[width=1\linewidth]{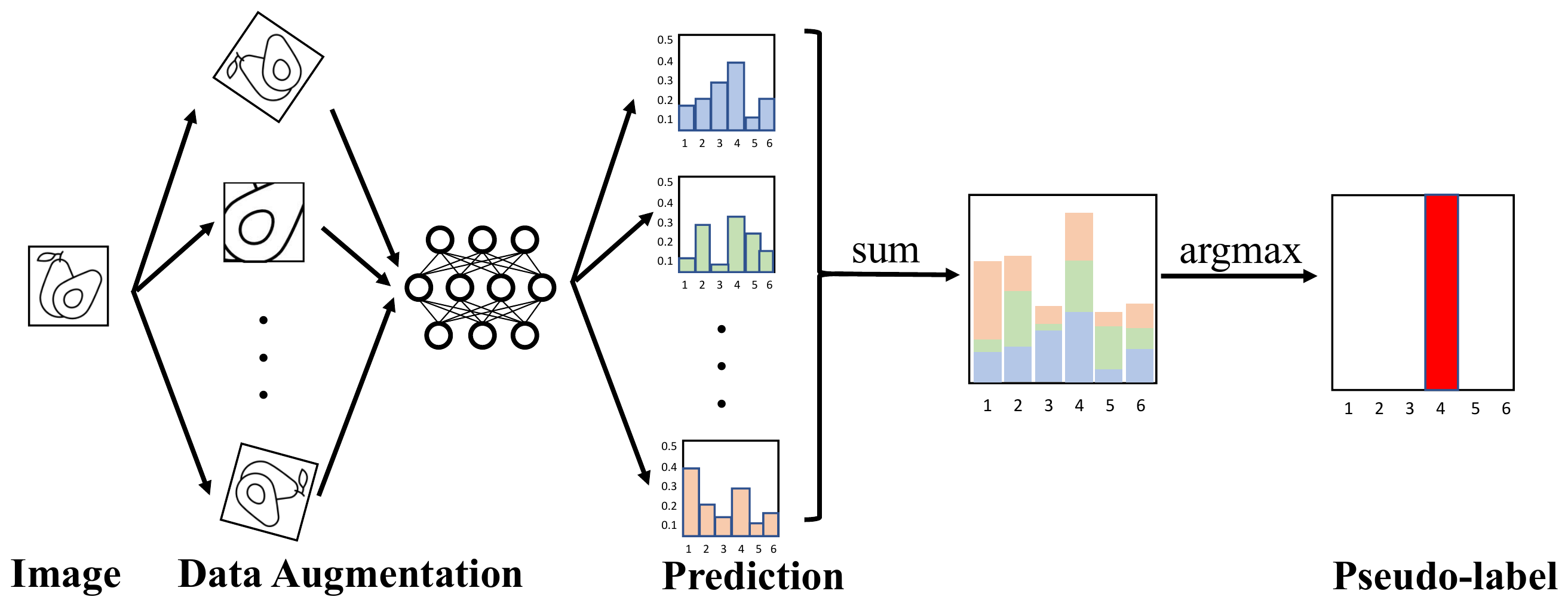}
	\caption{The overview of the proposed argmax method.}
	\label{fig-2}
\end{figure}

\subsubsection{Unsupervised Model Loss}
In SSL, consistent regularization and pseudo-label are commonly used methods for training models using unlabeled data.
Previous work naively combined these two methods without exploring the influence of the parameters in the unsupervised loss function on the model during the training process.
To this end, in our $\mathrm{FedMix}$ algorithm, we introduce dynamically adjusted hyperparameter items into the unsupervised loss function to alleviate the adverse effects caused by the parameters in the unsupervised loss. 
\textcolor{black}{We observe that the model has poor performance in the early stages of training, which resulted in a relatively insufficient quality of pseudo labels \cite{liu2020rc,long2020fedsemi}.
At this time, the method of pseudo-label will have a negative impact on training unsupervised model.
Consistency regularization method should account for the main contribution.
As the number of communication rounds increases, the performance of the unsupervised model gradually improves, and the pseudo labels will increasingly approach the ground-truth labels.
Thus, we should improve the status of pseudo-label method in training unsupervised model.
Moreover, in the middle and late stages of training, the unsupervised model is robust to unlabeled samples after data enhancement, so that the consistency regularization method has a negative impact.
Therefore, we need to appropriately increase the impact of pseudo-labels on the loss of unsupervised models, and reduce the impact of consistency regularization methods.}
Specifically, we define the following unsupervised model objective function:
\begin{equation}
\begin{aligned}
\mathop {\min }\limits_{\psi  \in {\mathbb{R}^d}} {{\cal L}_u}(\psi), {{\cal L}_u}(\psi) \buildrel\textstyle.\over= {\lambda _t}CE(\widehat y,{f_{\psi}}(u))+\\
{(1-\lambda _t)}||{f_{\psi}}({\pi _1}(u)) - {f_{\psi}}({\pi _2}(u))|{|^2}+ {\lambda _{L2}}||{\sigma} - \psi|{|^2},
\end{aligned}
\end{equation}
where $\lambda_{L2}$ is the regularization term to prevent the model from overfitting, ${u}$ is from unlabeled dataset $\mathcal{D}^u$, $\pi (\cdot)$ is the form of perturbation, i.e., $\pi_1$ is the shift augmentation, $\pi_2$ is the flip augmentation, $||\sigma - \psi||^2$ is a penalty term that aims to let the unsupervised model $\psi$ learn the knowledge of the supervised model $\sigma$, $\lambda_t$ is a hyperparameter that changes dynamically with training round $t$ as following:
\begin{equation}
{\lambda _t} = \frac{2}{\pi} \arctan \frac{FKt}{2BE}, {\lambda _t}\in (0,1),
\end{equation}
where $F$ represents the client participation rate, $K$ is the total number of clients, $t$ is the round of iteration, $B$ is the local min-batch size, and $E$ is the local training epoch. Furthermore, the $\hat{y}$ is pseudo label obtained by our proposed argmax method. As shown in Fig. \ref{fig-2}, the proposed argmax method is defined as follows:
\begin{equation}
\hat{y} = \mathsf{\textbf{1}} (\mathrm{Max} (\sum\limits_{i=1} ^{A} f_{\psi}(\pi_i (u)))),
\end{equation}
where $\mathrm{Max} (\cdot)$ is a function that can output the maximum probability that unlabeled data belongs to a certain class, $\mathsf{\textbf{1}} (\cdot)$ is the one-hot function that can change the numerical value to 1, $A$ represents the number of unlabeled data after data augmentation.
Since the argmax method adds additional computational overhead to the local client, therefore, limiting the number of pseudo-labels is the most direct way to reduce computational complexity \cite{9435947}.
\textcolor{black}{Active learning \cite{settles1995active,5272205} is a technique that systematically selects high-quality training samples from a data pool, which the goal is to use fewer data samples to achieve higher model performance.}
Thus, we utilize two active learning strategies to select specific unlabeled samples for pseudo-labeling and compare their corresponding model performance.
Based on information entropy, we can define the ``uncertainty" and ``min-entropy" selection strategy as follows:
\begin{equation}
\mathcal D_{n}^{uncertainty} = top_n [u \mapsto H(u)](D^{u}),
\end{equation}
\begin{equation}
\mathcal D_{n}^{min-entropy} = top_n [u \mapsto -H(u)](D^{u}),
\end{equation}
where $\mathcal D_{n}^{uncertainty}$ and $\mathcal D_{n}^{min-entropy}$ represent the dataset composed of the first $n$ unlabeled samples with large or small information entropy, respectively.
Because information entropy can reflect the uncertainty of unlabeled samples, the greater the entropy of unlabeled samples, the harder they are to identify.
Specifically, the calculation formula of information entropy is as follows:
\begin{equation}
H(u) = -\sum\limits_{j=1} ^{c}f_{\psi}(u)_j log(f_{\psi}(u)_j),
\end{equation}
where $c$ is denoted total class number, $f_{\psi}(u)_j$ denotes the value of belong to the $j$-th class of $f_{\psi}(u)$.
Furthermore, we also considered the strategy of randomly selecting $n$ unlabeled samples for pseudo-labeling in each round as comparative verification experiment.
\subsubsection{Supervised Model Loss}
Now we describe the classification loss of the supervised model learned from labeled data, i.e., the minimization of the supervised model objective function is defined as follows:
\begin{equation}
{\cal L}_s({\sigma}) \buildrel\textstyle.\over= {\lambda _s}CE(y,{f_{{\sigma}}}(x)),
\end{equation}
where $\lambda_s$ is the hyperparameter, ${x}$ and ${y}$ are from labeled dataset, and $f_{\sigma}({x})$ means the output of labeled samples on supervised model $\sigma$.
\subsubsection{Model Mixing}
In SSFL, the client uses the optimizer on the training set to update the local model and then uploads the model parameters or the updated gradient to the server. Finally, the server aggregates all client updates to obtain the next round of the global model.
However, the traditional method only focuses on updating the client model parameters, ignoring the implicit contribution of the global model in iteration.
In particular, under the non-IID data distribution, the global model may not be the optimal model because of the drift of the client model in the updating process.
Thus, in the model aggregation stage, we retain part of the global model information from the previous round to improve the robustness to non-IID data.
Therefore, the formal definition of the above strategy is as follows:
\begin{equation}\label{eq-6}
\omega_{t} = \alpha \psi_{t} + \beta \sigma_{t} + \gamma \omega_{t-1},
\end{equation}
where $\psi_{t}$ and $\sigma_{t}$ are the unsupervised model and the supervised model of the $t$-th round, the global model $\omega_{t-1}$ is from the previous round $t-1$, $\alpha$, $\beta$, and $\gamma$ are the corresponding weights of the three models (where $(\alpha ,\beta ,\gamma ) \in \{ \alpha  + \beta  + \gamma  = 1 \wedge \alpha ,\beta ,\gamma  \geqslant 0\}$).

\subsection{$\mathrm{FedFreq}$ Aggregation Algorithm}
The current FL framework follows a strategy of sampling clients to participate in training, that is, the server randomly selects only one group from all clients in each round to train the global model. Therefore, the number of times each client participates in training is unbalanced. In this section, we present the designed $\mathrm{FedFreq}$ aggregation algorithm, which can dynamically adjust the weight of the corresponding local model according to the training frequency of the client to alleviate the non-IID problem. We observe that the parameter distribution of the global model will be biased towards clients that often participate in federated training, which is not friendly to the robustness of the global model. Therefore, our insight is to reduce the influence of clients with high training frequency on the global model to improve the robustness of the model. Thus, the formal expression of the $\mathrm{FedFreq}$ aggregation algorithm is as follows:
\begin{equation}
\begin{aligned}
w_{t+1}^{k} = \dfrac{1 - p_{t+1}^{k}} 
{1 - p_{t+1}^{1}+ \cdots +1-p_{t+1}^{k}}
=\dfrac{1 - p_{t+1}^{k}}{FK-1},
\end{aligned}
\end{equation}
where $F$ is the sample proportion of the server, $K$ is the total number of clients, $p_{t+1}^{k} = \dfrac{q_{t+1}^{k}}{\sum\nolimits_{k \in \mathcal{K}_{t+1}}q_{t+1}^{k}}$, $q_{t+1}^{k}$ is the number of times that the $k$-th client has been trained up to the $t+1$-th round, and $\mathcal{K}_{t+1}$ denotes the set of clients selected by the server in round $t+1$.

\begin{figure*}[!t]
	\centering
	\large
	\subfigure [Labels-at-Client Scenario]{\includegraphics[width=0.45\linewidth]{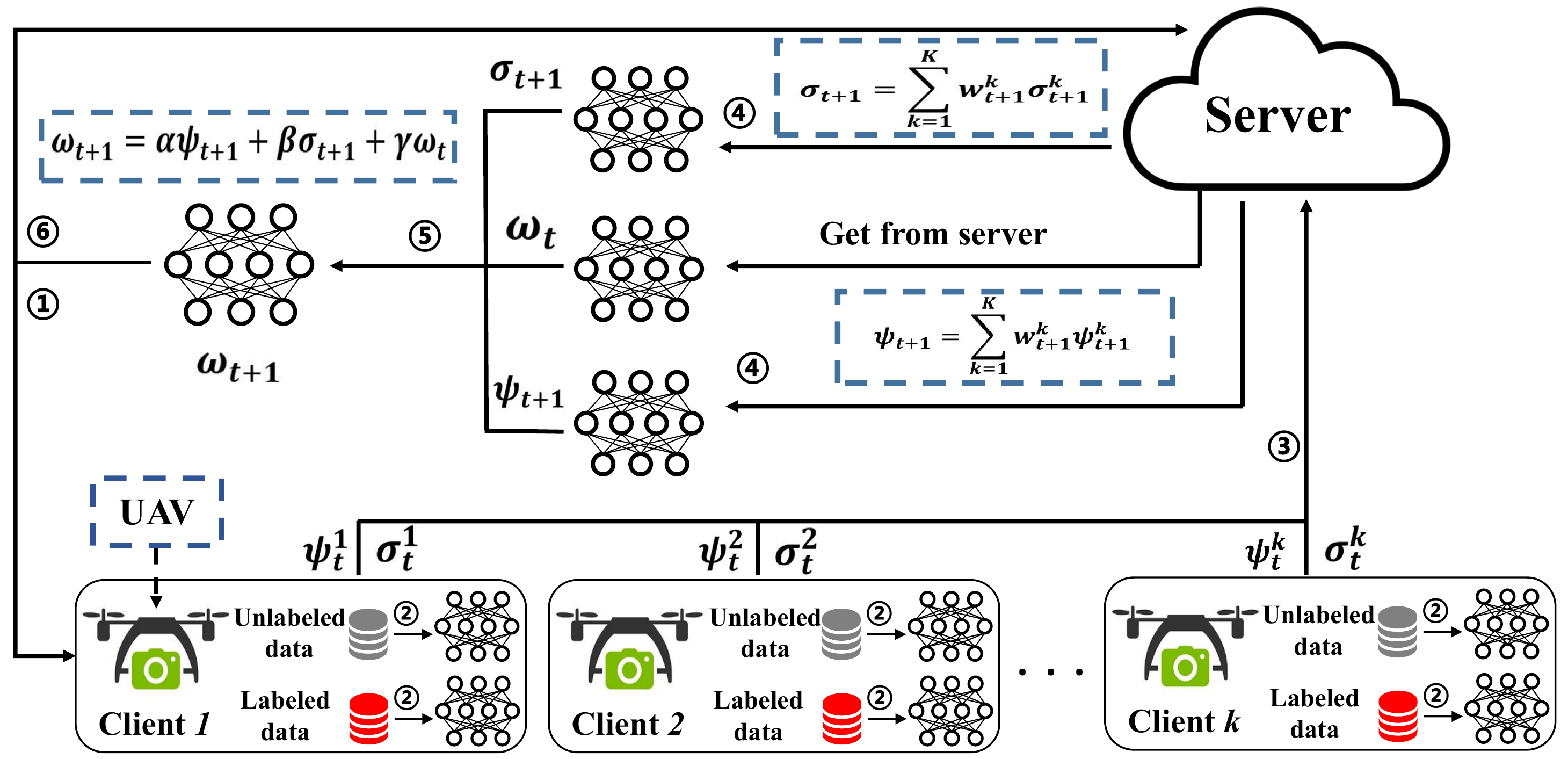}
		\label{fig-lc-1}}
	\hfill
	\subfigure[Labels-at-Server Scenario]{ \includegraphics[width=0.45\linewidth]{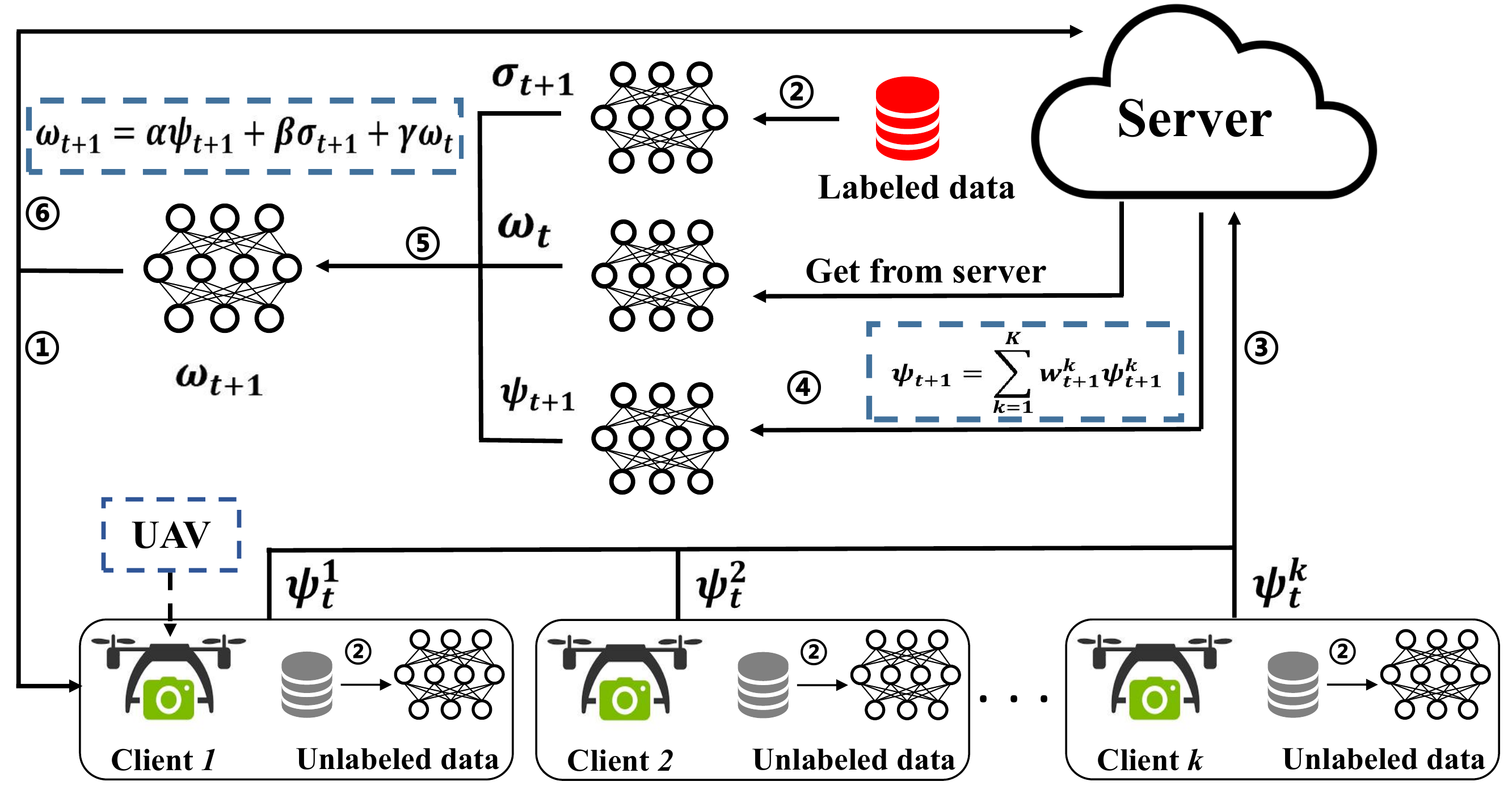}
		\label{fig-ls-1}}
\caption{\textcolor{black}{Overview of semi-supervised federated learning system in two scenarios.}}
	\label{fig-lc-ls}
\end{figure*}


\begin{algorithm}[!t]
	\caption{$\mathrm{FedMix}$ algorithm on labels-at-client.}
	\begin{algorithmic}[1]
	\REQUIRE The client set $\mathcal{K}$, $B_{u}$ is the local mini-batch size of client unlabeled data, $B_{s}$ is the mini-batch size of client labeled data, $E$ is the number of local epochs at the client side, and $\eta$ is the learning rate.
	\ENSURE The optimal global model $\omega^\ast$.
	\STATE \textbf{Server executes:}
	\STATE Initialize global model $\omega_0$
	\FOR{each round $t = 0, 1, 2, ...$}
		  \STATE $m \gets \mathrm{max}(F\cdot K, 1)$\\
		\STATE $S_t \gets$ randomly select $m$ clients from the client set $\mathcal{K}$\\
		  \FOR{each client $k \in S_t$ $\textbf{in parallel}$}
		  \STATE $\psi_{t+1}^{k}$, $\sigma_{t+1}^{k}$ $\gets$ \textbf{ClientUpdate($k, \omega_{t}^{k}$)}\\
		  \STATE $\psi_{t+1} = \sum\limits_{k=1} ^{K} w_{t+1}^{k} \psi_{t+1} ^{k}$  $// $Refer to $\mathrm{FedFreq}$ algorithm\\
		  \STATE $\sigma_{t+1} = \sum\limits_{k=1} ^{K} w_{t+1}^{k} \sigma_{t+1} ^{k}$  $// $Refer to $\mathrm{FedFreq}$ algorithm\\
		  \ENDFOR
		\STATE $\omega_{t+1} = \alpha \psi_{t+1} + \beta \sigma_{t+1} + \gamma \omega_{t}$\\
		\ENDFOR
		\STATE \textbf{ClientUpdate($k$, $\omega_{t}^{k}$):}$//$ Run on client $k$\\
		\STATE $\psi_{t}^k$ $\gets$ $\omega_{t}$
		\STATE $\sigma_{t}^k$ $\gets$ $\omega_{t}$
		\FOR{each local epoch $e$ from $1$ to $E$}
		    \FOR{minibatch $b_u \in B_{u}$ and $b_s \in B_{s}$} 
		    \STATE$\psi_{t+1}^{k}$ = $\psi_t^{k}$ - $\eta\bigtriangledown\mathcal{L}_{u}(k, \psi_{t}^{k}, \mathcal{D}_{k}^{u}, b_u)$\\
		    \STATE$\sigma_{t+1}^{k}$ = $\sigma_t^{k}$ - $\eta\bigtriangledown\mathcal{L}_{s}(k, \sigma_{t}^{k}, \mathcal{D}_{k}^{s}, b_s)$\\
		    \ENDFOR
		  \ENDFOR
			\RETURN $\psi_{t+1}^{k}$, $\sigma_{t+1}^{k}$ to server.
		\end{algorithmic}
	\end{algorithm}

\subsection{$\mathrm{FedMix}$ Algorithm for Two Practical Scenarios}\label{sec-6}
\subsubsection{Labels-at-Client Scenario}
In the labels-at-client scenario, local clients have a small amount of labeled data and many unlabeled data, while the server has no data. The training steps are represented in Fig. \ref{fig-lc-1} above. The overall learning process of the global model is similar to traditional FL, except that the local client needs to train both the supervised model and the unsupervised model. In this case, the FL model training is iterated according to the following protocol:

\begin{itemize}
  \item [1)] 
  The server randomly selects a certain proportion of $F$ ($0<F<1$) clients from all local clients to send the initialized global model $\omega_{0} $. 
  \item [2)]
  The selected clients perform SGD training on their local labeled and unlabeled datasets to update their local supervised and unsupervised models.
  \item [3)]
  The server collects the models of the selected clients and uses the $\mathrm{FedFreq}$ (see Section IV-B) aggregation algorithm to obtain the global supervised model $\sigma$ and the unsupervised model $\psi$. Then the server mixes the unsupervised global model, supervised model, and the previous round of global model according to different weights to get a new round of global model, i.e., Equation \eqref{eq-6}.
\end{itemize}

Similar to FL, the client uses unlabeled data and labeled data to train the local supervised model and the unsupervised model, and the server only aggregates model parameters. The training details are given in Algorithm 1.


\subsubsection{Labels-at-Server Scenario}
We now describe another more challenging scenario in SSFL, namely the labels-at-server scenario, in which the labeled data is located on the server side, and the unlabeled data is only available on the client side, as shown in Fig. \ref{fig-ls-1}.
The iterative process of the entire system is as follows:

\begin{itemize}
  \item [1)] 
  The server randomly selects a certain proportion of clients to deliver the initialized model parameter $\omega_o$. Additionally, $\omega_o$ is also stored on the server side for supervised learning training.   
  \item [2)]
   The server uses the local optimizer on the labeled dataset $\mathcal{D}_s$ to train the supervised model $\sigma$ (i.e., $ \sigma_t \leftarrow {\omega _t}$). Meanwhile, for the $k$-th client, it utilizes the local unlabeled dataset to train the received global model $\omega_{t}$ (i.e., $ \psi_t^k \leftarrow {\omega _t}$) and then obtains the unsupervised model $\psi_{t+1} ^{k}$.
  \item [3)]
  The server uses the proposed $\mathrm{FedFreq}$ aggregation algorithm to aggregate the unsupervised models uploaded by the clients to obtain the global unsupervised model, i.e.,$\psi_{t+1} = \sum\limits_{k=1} ^{K} w_{t+1}^{k} \psi_{t+1} ^{k},$ where $\psi_{t+1} ^{k}$ is the unsupervised model of the $k$-th client at $t+1$-th training round and $w_{t+1}^{k}$ is the weight of the $k$-th client. The server then aggregates the global unsupervised model $\psi_{t+1}$, the supervised model $\sigma_{t+1}$, and the global model $\omega_{t}$ from the previous round $t$ to obtain a new global model $\omega_{t+1}$, i.e., Equation \eqref{eq-6}.
\end{itemize}

Note that unlike FL, in labels-at-server scenario, the server not only aggregates the model uploaded by the clients, but also trains the supervised model $\sigma$ on the labeled dataset $\mathcal{D}_s$. More training details are described in Algorithm 2.

\begin{algorithm}[!t]
	\caption{$\mathrm{FedMix}$ algorithm on labels-at-server.}
	\begin{algorithmic}[1]
	\REQUIRE The client set $\mathcal{K}$, $B_{s}$ is the mini-batch size at the server side, $E_{s}$ is the number of epochs at the server side,
		$B_{u}$ is the local mini-batch size at the client side,
		$E_{u}$ is the number of local epochs at the client side,
		and $\eta$ is the learning rate.
	\ENSURE The optimal global model $\omega^\ast$.
	\STATE \textbf{Server executes:}
	\STATE Initialize global model $\omega_0$
	\FOR{each round $t = 0, 1, 2, ...$}
	\STATE $\sigma_{t}$ $\gets$ $\omega_{t}$
		  \FOR{the server epoch $e$ from $1$ to $E_{s}$}
		    \FOR{mini-batch $b \in B_{s}$}
		    \STATE $\sigma_{t+1}$ = $\sigma_t$ - $\eta\bigtriangledown\mathcal{L}_{s}(\sigma_t, \mathcal{D}_s, b)$\\
		    \ENDFOR
		  \ENDFOR
		  \STATE $m \gets \mathrm{max}(F\cdot K, 1)$\\
		\STATE $S_t \gets$ randomly select $m$ clients from the client set $\mathcal{K}$\\
		  \FOR{each client $k \in S_t$ $\textbf{in parallel}$}
		  \STATE $\psi_{t}^k$ $\gets$ $\omega_{t}$
		  \STATE $\psi_{t+1}^{k} \gets$ \textbf{ClientUpdate($k, \psi_{t}^{k}$)}\\
		  \ENDFOR
		\STATE $\psi_{t+1} = \sum\limits_{k=1} ^{K} w_{t+1}^{k} \psi_{t+1} ^{k}$  $// $ Refer to $\mathrm{FedFreq}$ algorithm\\
		\STATE$\omega_{t+1} = \alpha \psi_{t+1} + \beta \sigma_{t+1} + \gamma \omega_{t}$\\
		\ENDFOR
		\STATE \textbf{ClientUpdate($k$, $\psi_{t}^{k}$):}$//$ Run on client $k$\\
		\FOR{each local epoch $e$ from $1$ to $E_{u}$}
		    \FOR{minibatch $b \in B_{u}$}
		    \STATE$\psi_{t+1}^{k}$ = $\psi_t^{k}$ - $\eta\bigtriangledown\mathcal{L}_{u}(k, \psi_{t}^{k}, \mathcal{D}_{k}^{u}, b)$\\
		    \ENDFOR
		  \ENDFOR
			\RETURN $\omega^\ast$ to server.
		\end{algorithmic}
	\end{algorithm}

\section{Experiment}\label{sec-7}
In the labels-at-client and labels-at-server scenario, we compare baselines with three different tasks on two datasets to experimentally validate our method $\mathrm{FedMix}$. For the two real-world datasets (i.e., CIFAR-10 and Fashion-MNIST), we simulate the FL setup (one server and $K$ clients) on a commodity machine with Intel(R) Core(TM) i9-9900K CPU @ 3.60GHz and NVIDIA GeForce RTX 2080Ti GPU.
\subsection{Experiment Setup}
\textbf{Dataset:} 
\textcolor{black}{In our experiment, the CIFAR-10 and Fashion-MNIST datasets are used to simulate SSFL-based UAV image classification tasks.
The detailed settings of these two datasets in the two SSFL scenarios can be obtained from Table \ref{dataset}.
For the streaming setting of the Fashion-MNIST dataset, the local data of each client is equally divided into 10 parts.
In the process of training, only one part of the data is used in each round. 
Furthermore, to simulate the setting of non-IID, we introduced the Dirichlet distribution function.
Specifically, we generate data distributions of different non-IID levels by adjusting the parameters of the Dirichlet distribution function (i.e., $ \mu $).
As shown in Fig. 4, the smaller the $\mu $, the higher the non-IID level of the data distribution of each client; otherwise, the data distribution of the client tends to the IID setting.}

\begin{table}[!t]
\centering
\caption{\textcolor{black}{The settings of CIFAR-10 and Fashin-MNIST datasets in two scenarios of SSFL.}}
\label{dataset}
\begin{tabular}{|c|c|c|c|}
\hline
\multicolumn{4}{|c|}{CIFAR-10}                                                                                                      \\ \hline
\multicolumn{1}{|c|}{\multirow{2}{*}{Case}} & \multicolumn{2}{c|}{Training set}                                    & \multirow{2}{*}{Test set} \\ \cline{2-3}
\multicolumn{1}{|c|}{}                      & \multicolumn{1}{c|}{labeled} & \multicolumn{1}{c|}{unlabeled} &                       \\ \hline
\multicolumn{1}{|c|}{Labels-at-Client}      & \multicolumn{1}{c|}{5000}    & \multicolumn{1}{c|}{50000}     & \multirow{2}{*}{2000} \\ \cline{1-3}
\multicolumn{1}{|c|}{Labels-at-Server}      & \multicolumn{1}{c|}{1000}    & \multicolumn{1}{c|}{55000}     &                       \\ \hline
\multicolumn{4}{|c|}{Fashion-MNIST}                                                                                                 \\ \hline
\multicolumn{1}{|c|}{Labels-at-Client}      & \multicolumn{1}{c|}{5000}    & \multicolumn{1}{c|}{58000}     & \multirow{2}{*}{2000} \\ \cline{1-3}
\multicolumn{1}{|c|}{Labels-at-Server}      & \multicolumn{1}{c|}{1000}    & \multicolumn{1}{c|}{63000}     &                       \\ \hline
\end{tabular}
\end{table}

\begin{figure}[t]
	\centering
	\subfigure []{\includegraphics[width=0.45\linewidth]{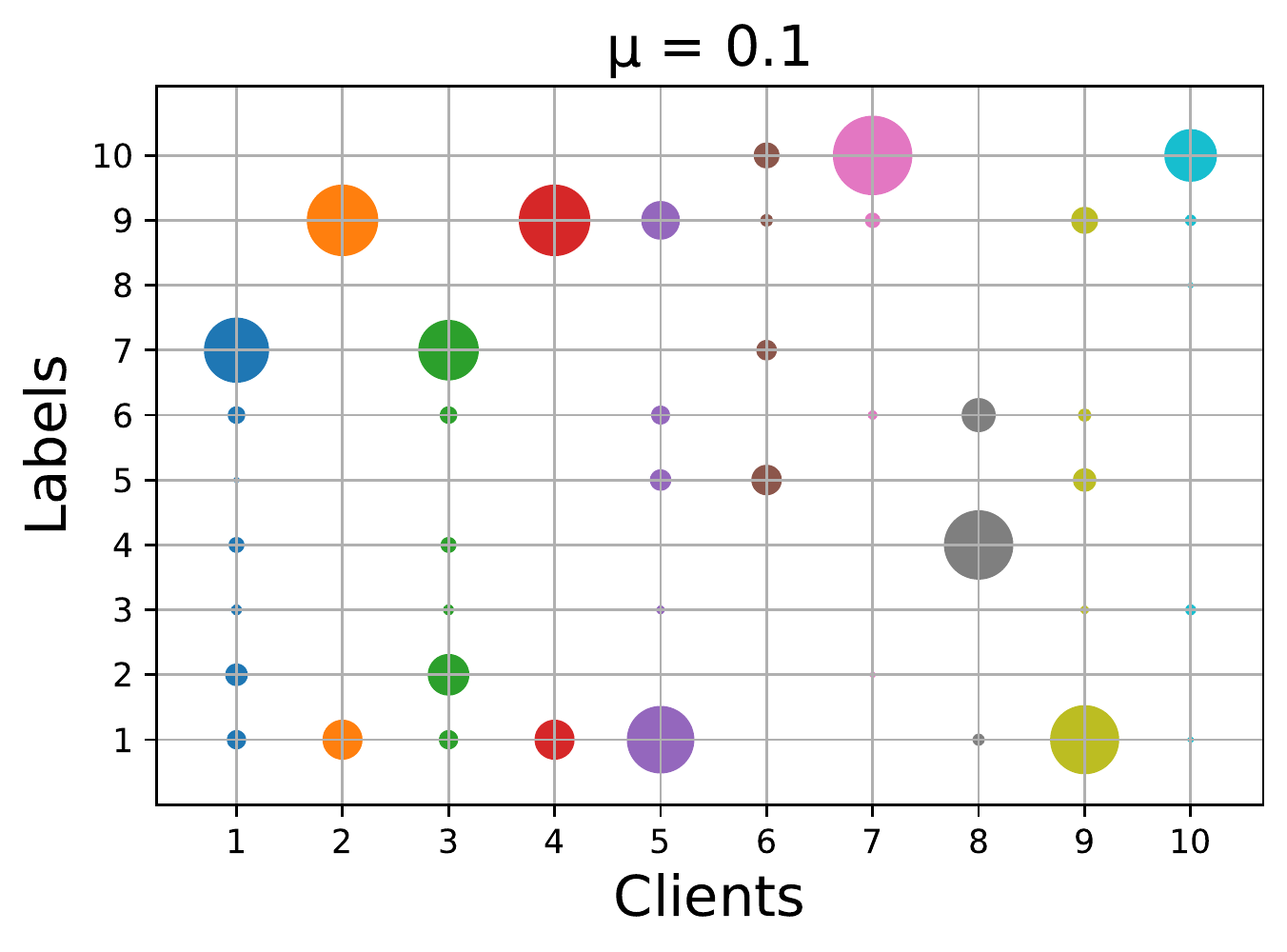}
		\label{a-11}}
	\subfigure[]{\includegraphics[width=0.45\linewidth]{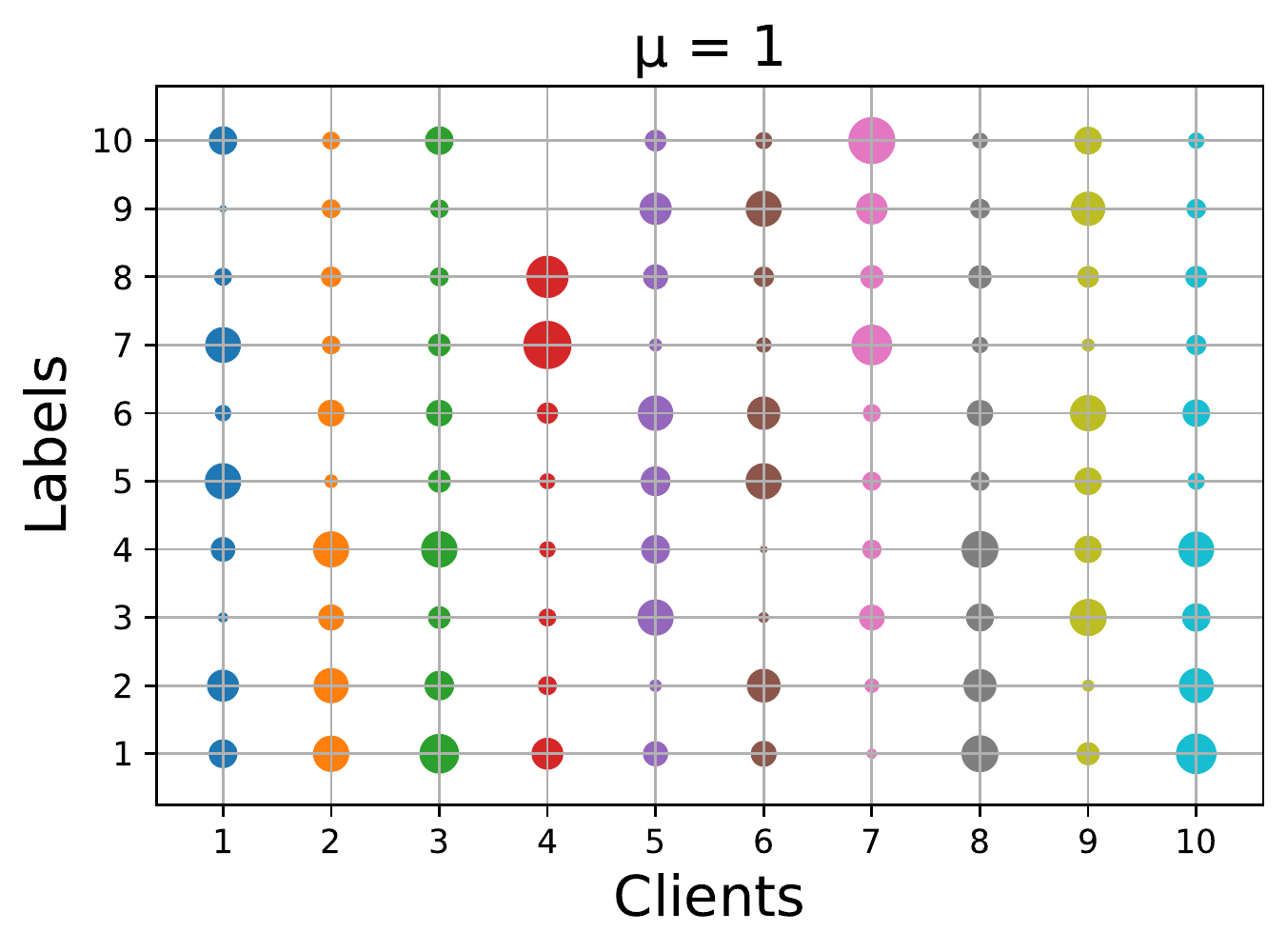}
		\label{b-12}}
	\caption{The non-IID levels of 10 clients are on a dataset with 10 classes, figure adapted from \cite{hsu2019measuring}.}
	\label{fig-3}
\end{figure}

\begin{table*}[!t]
	\centering
	\caption{Hyperparameters setting details}
	\begin{tabular}{|c|c|c|c|c|c|c|c|c|c|c|c|c|c|}\hline
		\multicolumn{14}{|c|}{Labels-at-Client}\\\hline
		Methods & $t$ & $\eta$ & $\lambda_{s}$ & $\lambda_{u}$ & $\lambda_{t}$ & $\lambda_{L2}$ & $\alpha$ & $\beta$ & $\gamma$ & $E_{u}$ & $E_{s}$ & $B_{u}$ & $B_{s}$\\\hline
		SL-FedAvg & 600 & 1e-2 & 10 & - & - & - & - & - & - & - & 1 & - & 10\\
		SSL-FedAvg & 600 & 1e-2 & 10 & 1 & - & - & - & - & - & 1 & 1 & 100 & 10\\
		FedMatch & 600 & 1e-2 & 10 & - & 1 & 10 & 1 & 1 & 0 & 1 & 1 & 100 & 10\\
		\textbf{FedMix-FedAvg} & 600 & 1e-2 & 10 & - & (0, 1) & 15 & 0.5 & 0.3 & 0.2 & 1 & 1 & 100 & 10\\
		\textbf{FedMix-FedFreq} & 600 & 1e-2 & 10 & - & (0, 1) & 15 & 0.5 & 0.3 & 0.2 & 1 & 1 & 100 & 10\\\hline
		\multicolumn{14}{|c|}{Labels-at-Server}\\\hline
		SL-FedAvg & 150 & 1e-3 & 10 & -  & - & - & - & - & - & - & 1 & - & 64\\
		SSL-FedAvg & 150 & 1e-3 & 10 & 1 & - & - & - & - & - & 1 & 1 & 100 & 64\\
		FedMatch & 150 & 1e-3 & 10 & - & 1 & 10 & 1 & 1 & 0 & 1 & 1 & 100 & 64\\
		\textbf{FedMix-FedAvg} & 150 & 1e-3 & 10 & - & (0, 1) & 15 & 0.5 & 0.3 & 0.2 & 1 & 1 & 100 & 64\\
		\textbf{FedMix-FedFreq} & 150 & 1e-3 & 10 & - & (0, 1) & 15 & 0.5 & 0.3 & 0.2 & 1 & 1 & 100 & 64\\\hline
	\end{tabular}
	\label{tab-canshu1}
\end{table*}
\textbf{Baseline:} Our baselines are: 1) $\mathrm{SL-FedAvg}$: federated learning for supervised training using sufficient labeled samples.
2) $\mathrm{SSL-FedAvg}$: standard semi-supervised federated learning with a naive combination of consistent regularization and pseudo-label methods under limited labeled samples and a large number of unlabeled samples.
3) $\mathrm{FedMatch}$ \cite{jeong2021federated} : semi-supervised federated learning  naively using unsupervised model and supervised model parameter decomposition strategy (i.e., $\omega = \psi + \sigma$) under limited labeled samples and massive unlabeled samples.

\textbf{Training details:} In the training process, our model and baseline use Stochastic Gradient Descent (SGD) to optimize the ResNet-9 neural network with initial learning rate $\eta=1e-2$ or $\eta=1e-3$. In different scenarios, we set training round $t=600$ or $150$, the unsupervised learning training epoch $E_{u}=1$ and mini-batch size $B_{u}=100$, the supervised learning training epoch $E_{s}=1$ and mini-batch size $B_{s}= 10$ or $64$. Moreover, we set the data augmentation number in the argmax method $A = 3$ or $5$, $n = 100$ in active learning strategy, the number of labeled samples on sever is $N_s=1000$, the number of labeled samples on each client is  $N_k = 50$. Refer to Table \ref{tab-canshu1} for detailed parameter settings.

Finally, we ensure that all hyperparameters are set reasonably for $\mathrm{FedMatch}$ and our method for fair evaluation and comparison. For all experiments, due to the uncontrollable randomness of the training, we give the average accuracy and fluctuation range of the three training results, as shown in Tables \ref{tab-1}, \ref{tab-2} and \ref{tab-3}.
\begin{table}[!t]
	\centering
	\caption{Performance comparison of different methods in scenario labels-at-client.}
	\begin{tabular}{|c|c|c|}\hline
		\multicolumn{3}{|c|}{CIFAR-10 with 100 clients (K=100, F=0.05, A=3)}\\\hline
		Labels-at-Client&\multicolumn{2}{c|}{Accuarcy(\%)}\\\hline
		Methods&IID&non-IID\\\hline
		SL-FedAvg&81.25$\pm$0.11&78.40$\pm$0.32\\
		SSL-FedAvg &46.53$\pm$0.21&43.65$\pm$0.54\\
		FedMatch&51.64$\pm$0.24&51.24$\pm$0.47\\
		FedMix-FedAvg&62.05$\pm$0.18&62.17$\pm$0.32\\
		\textbf{FedMix-FedFreq}&\textbf{63.39$\pm$0.17}&\textbf{62.78$\pm$0.26}\\\hline
	\end{tabular}
	\label{tab-1}
\end{table}

\begin{figure}[!t]
	\centering
	\large
	\subfigure [Labels-at-Client Scenario]{\includegraphics[width=0.45\linewidth]{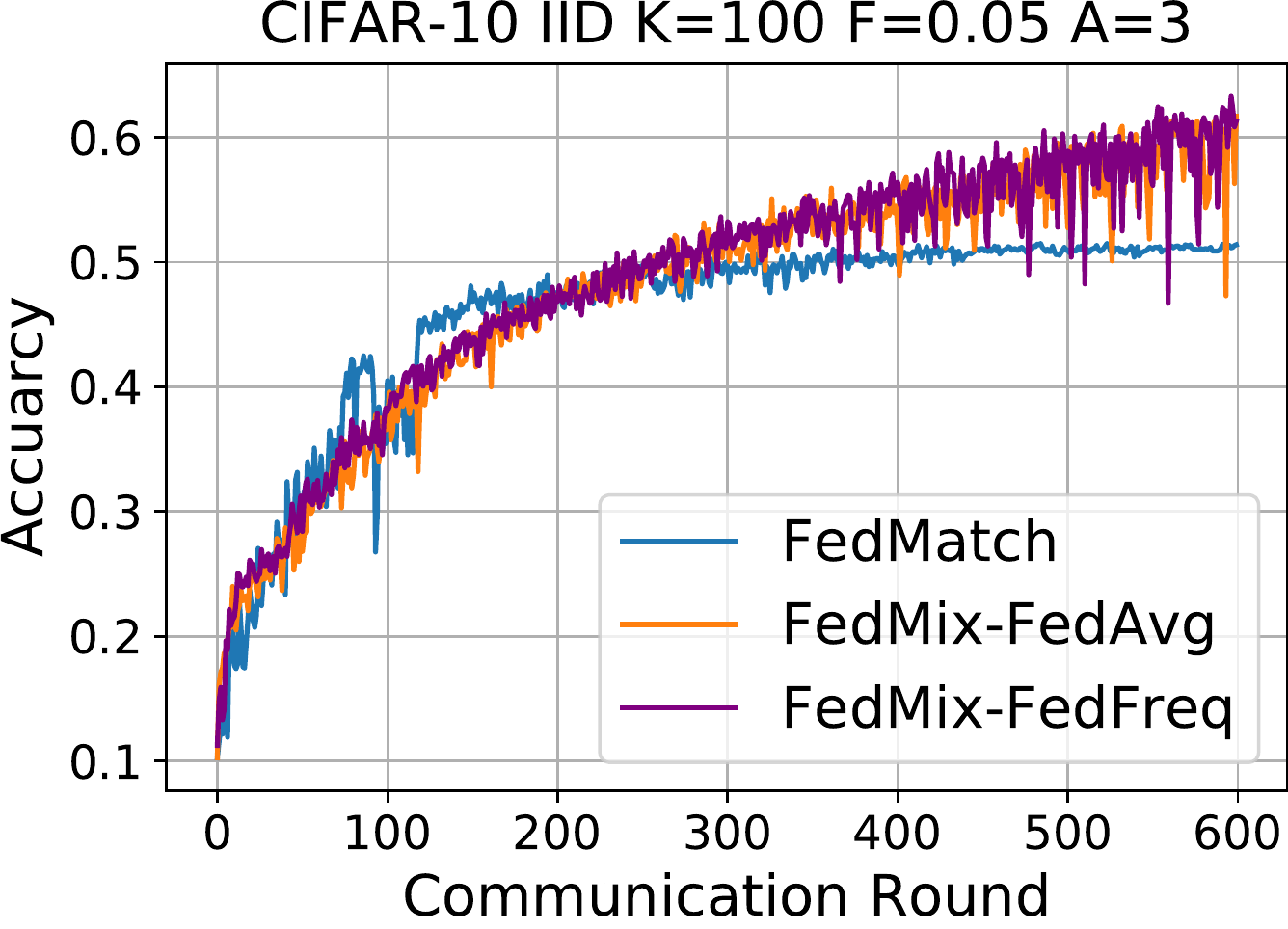}
		\label{lc-21}}
	\hfill
	\subfigure[Labels-at-Client Scenario]{	\includegraphics[width=0.47\linewidth]{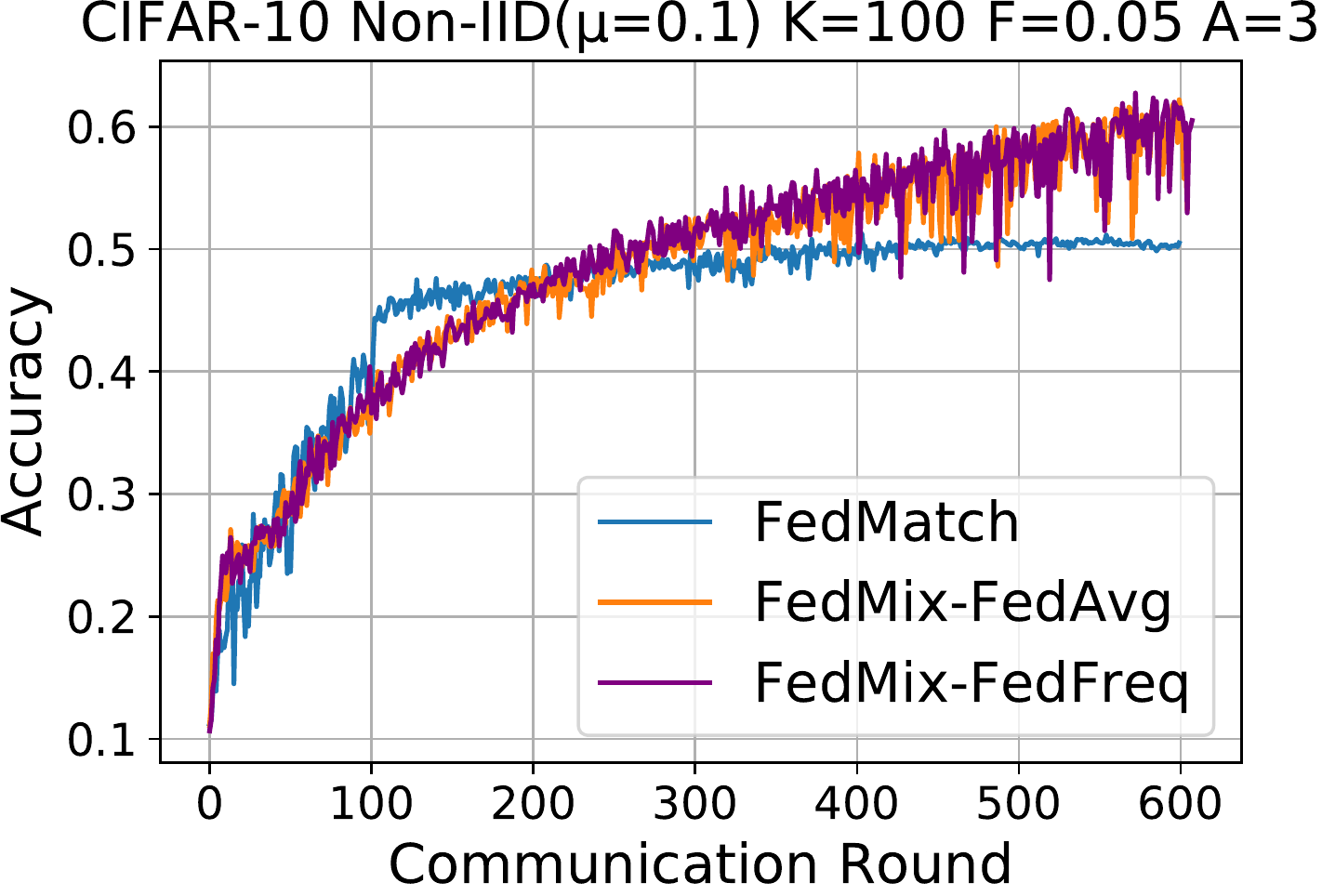}
		\label{lc-22}}
	\caption{Test accuracy curves of IID and non-IID in labels-at-client scenario.}
	\label{fig-lc-2}
\end{figure}

\subsection{Experiment Results}
\subsubsection{Performance evaluation of CIFAR-10 dataset IID and non-IID settings in labels-at-client scenario} Under IID and non-IID settings, we can observe in Table \ref{tab-1} that with the improvement of the SSFL method, its performance gradually approaches $\mathrm{SL-FedAvg}$, which indicates that when the number of labeled data is limited, unlabeled data plays a vital role in improving model performance. Meanwhile, it also proves that the semi-supervised learning method is well applied under the framework of federated learning. Moreover, our approach is superior to $\mathrm{FedMatch}$ using naive parameter decomposition in both settings. In particular, the $\mathrm{FedFreq}$ aggregation rule improves the accuracy by about 1\% compared to $\mathrm{FedAvg}$.

Intuitively, from Fig. \ref{lc-21} and Fig. \ref{lc-22}, we can observe that our proposed method significantly outperforms the baseline performance under IID and non-IID settings. Specifically, the performance of our model is gradually improving with the increase of training rounds, reaching an accuracy of 63\%, while the baseline finally converges to 51.5\%. This is because the baseline method ignores the implicit contribution of the global model in the iterative process, in which performance converges rapidly after a certain number of communication rounds.

\begin{table}[!t]
	\centering
	\caption{Performance comparison of different methods in scenario labels-at-server.}
	\begin{tabular}{|c|c|c|}\hline
		\multicolumn{3}{|c|}{CIFAR-10 with 100 clients (K=100, F=0.05, A=5)}\\\hline
		Labels-at-Server&\multicolumn{2}{c|}{Accuarcy(\%)}\\\hline
		Methods&IID&non-IID\\\hline
		SL-FedAvg&N/A&N/A\\
		SSL-FedAvg&24.56$\pm$0.38&27.45$\pm$0.83\\
		FedMatch&44.56$\pm$0.23&46.44$\pm$0.35\\
		FedMix-FedAvg&47.10$\pm$0.19&46.05$\pm$0.72\\
		\textbf{FedMix-FedFreq}&\textbf{47.58$\pm$0.14}&\textbf{47.92$\pm$0.17}\\\hline
	\end{tabular}
	\label{tab-2}
\end{table}

\begin{figure}[!t]
	\centering
	\large
	\subfigure [Labels-at-Server Scenario]{\includegraphics[width=0.43\linewidth]{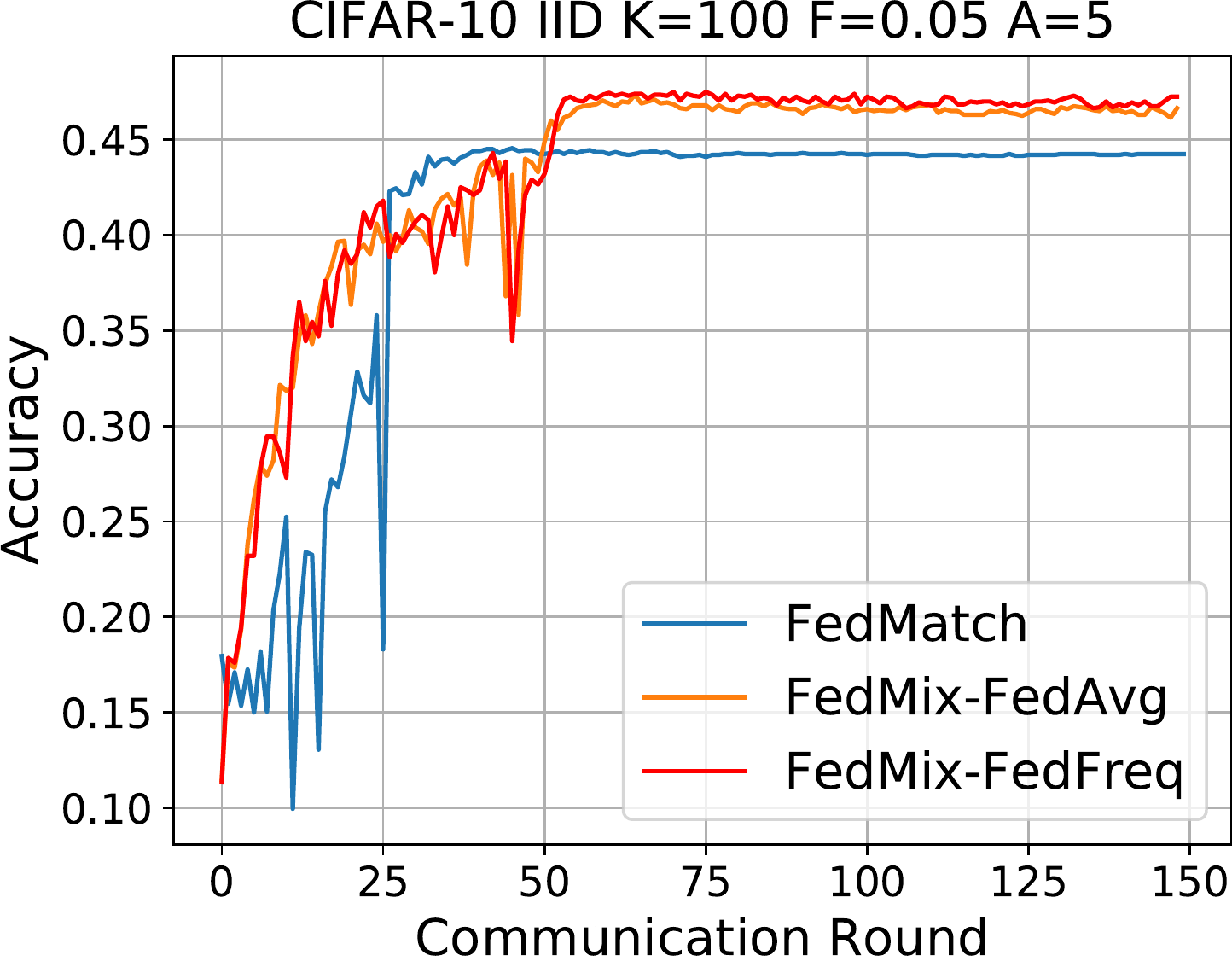}
		\label{ls-21}}
	\hfill
	\subfigure[Labels-at-Server Scenario]{	\includegraphics[width=0.43\linewidth]{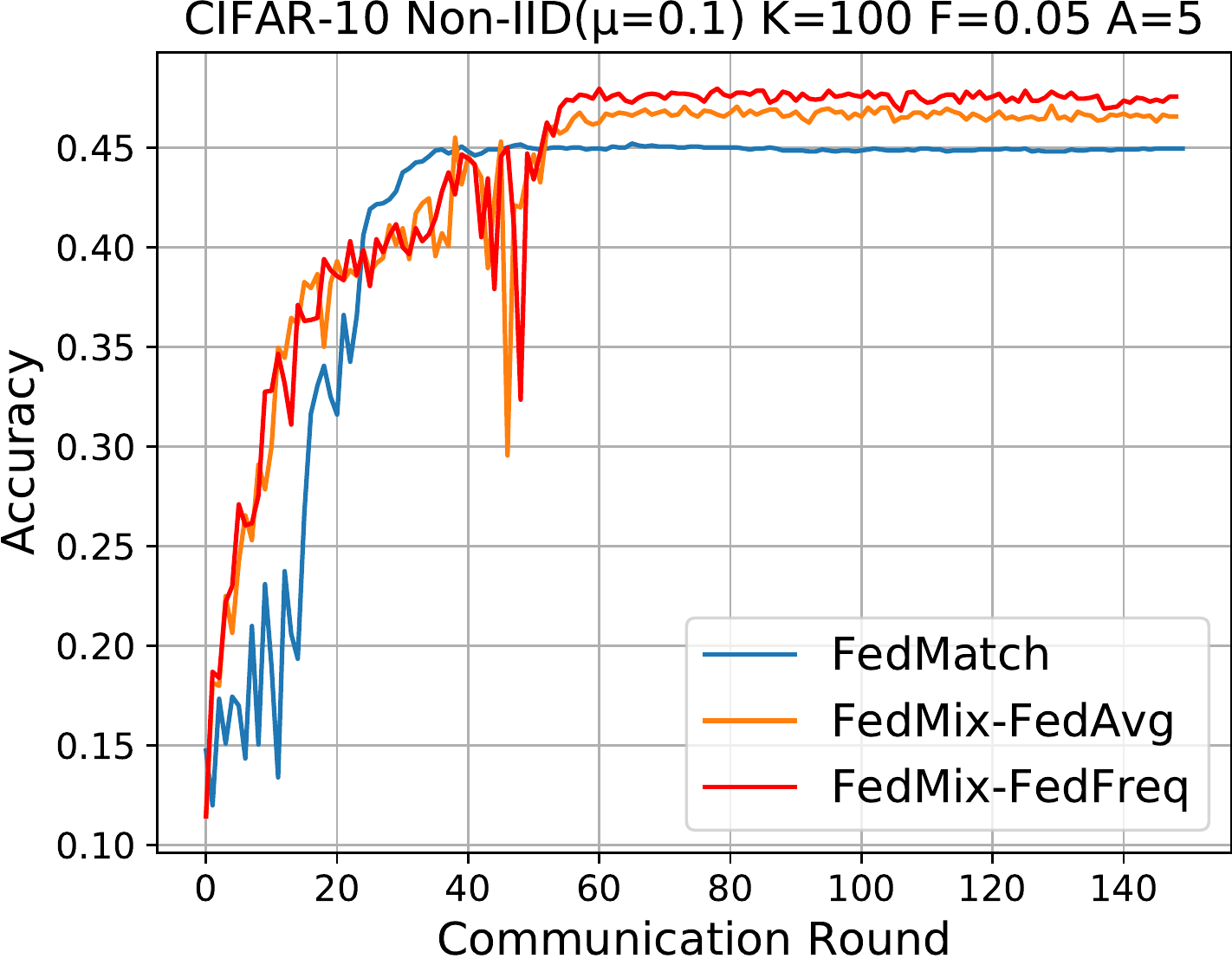}
		\label{ls-22}}
	\caption{Test accuracy curves of IID and non-IID in labels-at-server scenario.}
	\label{fig-ls-2}
\end{figure}

\subsubsection{Performance evaluation of CIFAR-10 dataset IID and non-IID settings in labels-at-server scenario} The labels-at-server scenario is more challenging than the labels-at-client scenario. As can be seen from Table \ref{tab-2}, in this scenario, the naive combination of federated learning and semi-supervised learning has the problem of knowledge forgetting. The knowledge learned by the model from labeled data is easily disturbed by the task of unlabeled data. The $\mathrm{FedMatch}$ uses model parameter decomposition to solve this problem effectively. Significantly, we further improve the performance by about 3\% by observing the implicit contribution of the global model in the iteration.

As shown in Fig. \ref{fig-ls-2}, IID and non-IID settings for the CIFAR-10 dataset, our method $\mathrm{FedMix}$ is better than baseline under each different aggregation method settings. For example, under the non-IID setting, the convergence accuracy of our method is 47.5\% about 3\% higher than that of the baseline. In particular, the accuracy of our method increases faster and more stable in the early stage of model training. The reason is that: (1) The $\mathrm{FedMix}$ focuses on the implicit contributions between iterations of the global model in a fine-grained manner, while the $\mathrm{FedMatch}$ only
naively uses model parameter decomposition. (2) Frequency-based aggregation method $\mathrm{FedFreq}$ is more suitable for non-IID settings.

\begin{figure}[!t]
	\centering
	\large
	\subfigure [Labels-at-Client Scenario]{\includegraphics[width=0.45\linewidth]{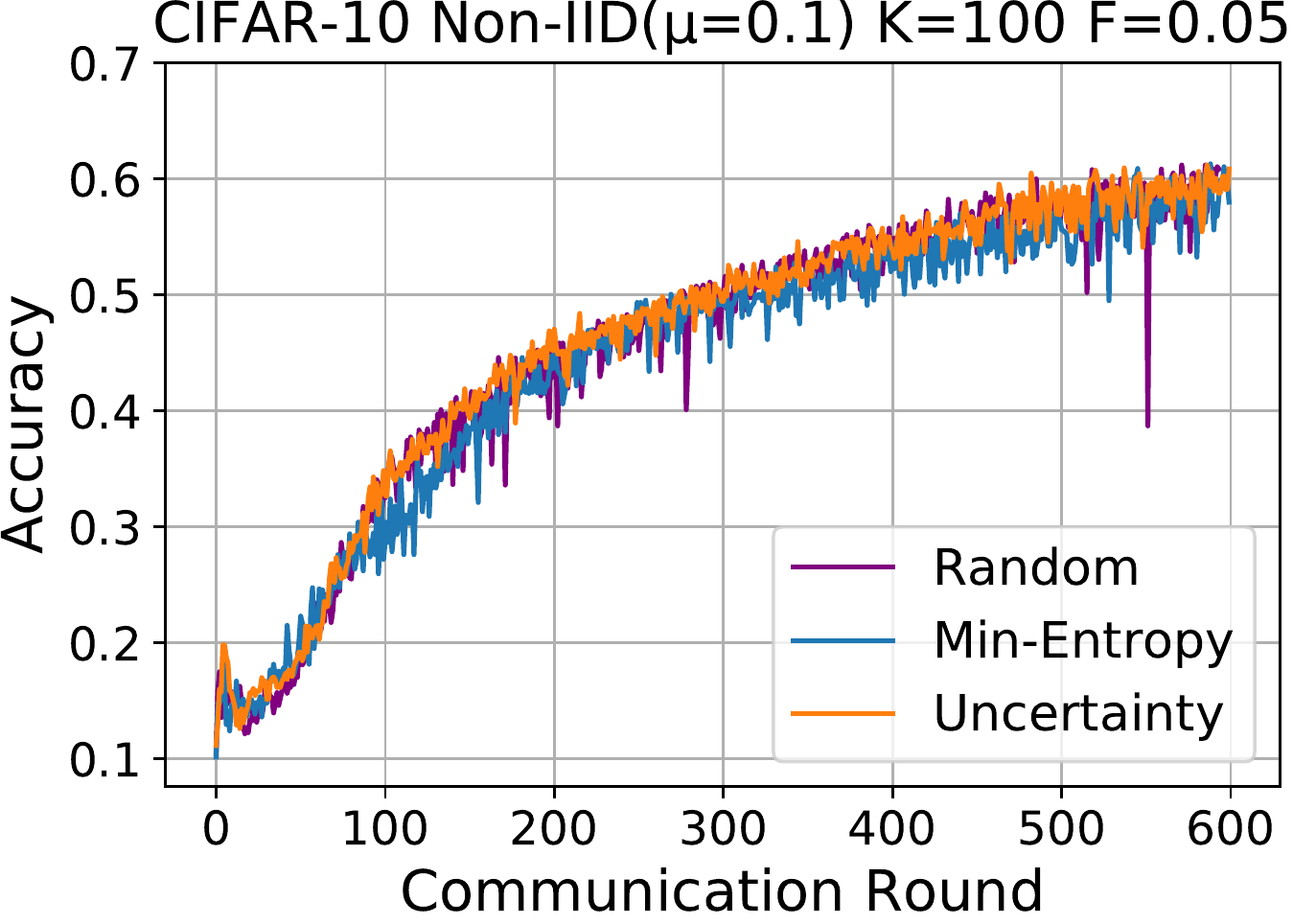}
		\label{lc-31}}
	\hfill
	\subfigure[Labels-at-Client Scenario]{	\includegraphics[width=0.45\linewidth]{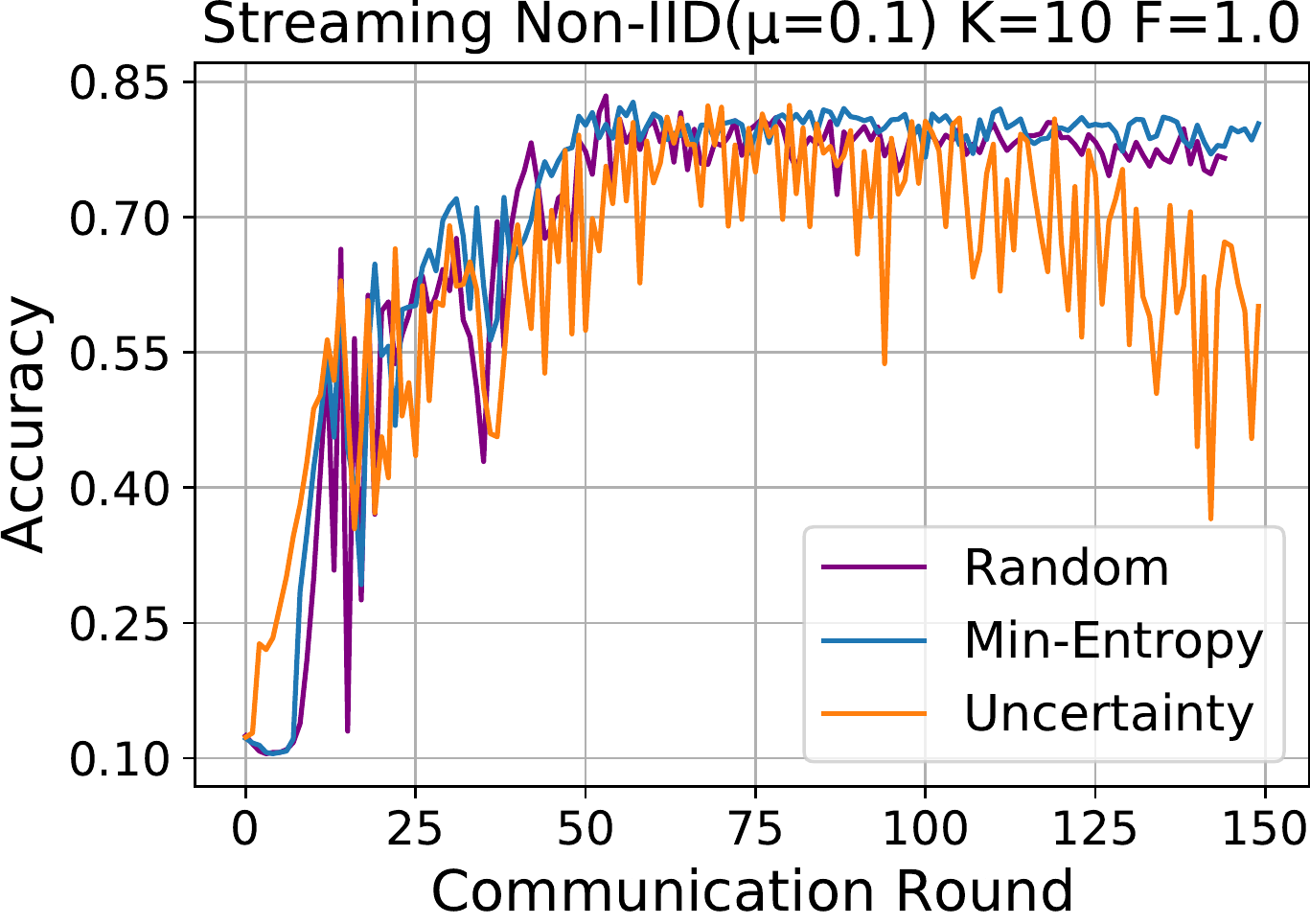}
		\label{lc-32}}

	\subfigure [Labels-at-Server Scenario]{\includegraphics[width=0.45\linewidth]{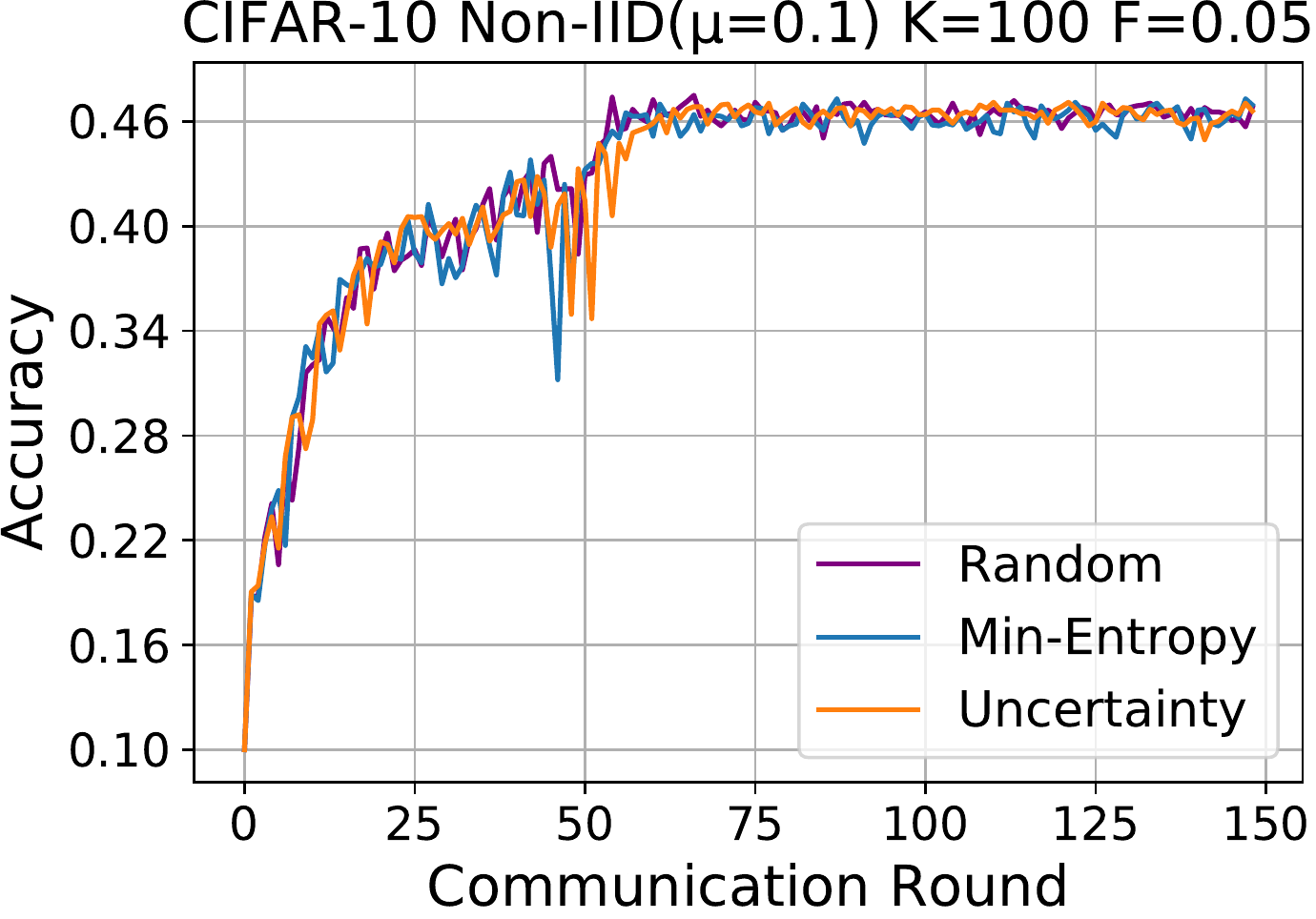}
		\label{ls-31}}
	\hfill
	\subfigure[Labels-at-Server Scenario]{	\includegraphics[width=0.45\linewidth]{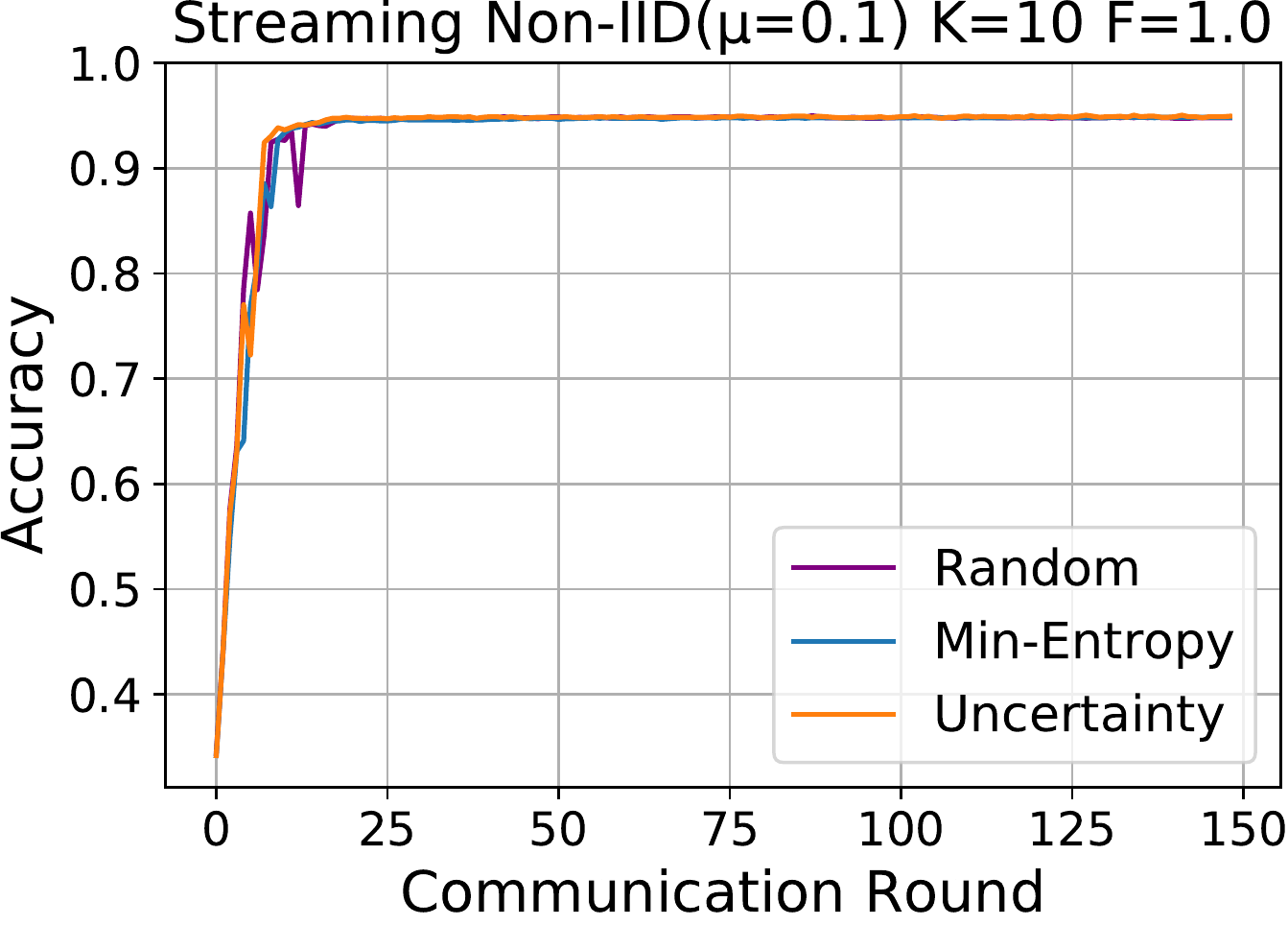}
		\label{ls-32}}
	\textcolor{black}{\caption{Performance comparison of different selection strategies for CIFAR-10 and Fashion-MNIST datasets in two scenarios.}}
	\label{ls-fig-3}
\end{figure}

\subsubsection{Performance evaluation of different selection strategies in two scenarios}
Fig. \ref{lc-31} shows the model performance comparison of different selection strategies on the CIFAR-10 dataset in label-at-client scenario. The three selection strategies can steadily improve the performance of the model, and the uncertain strategy performs slightly better. 
This is because the pseudo-label is generated by the local model combined with data augmentation methods. When the accuracy of the local model is low, the min-entropy strategy will significantly reduce the quality of pseudo-labels. However, uncertain samples have more information, which can enable the model to learn more useful knowledge. Thus, in this scenario and CIFAR-10 dataset, we adopt an uncertain active learning strategy to pseudo-label unlabeled samples.

On the other hand, for the Fashion-MNIST dataset, we can see from Fig. \ref{lc-32} that the three selection strategies show great differences, especially the uncertain active learning selection strategy leads to the problem of performance degradation.
In the Fashion-MNIST dataset, our model cannot correctly classify unlabeled samples with large information entropy, resulting in low-quality pseudo-labels. Meanwhile, as the number of iterations increases, the weight of the pseudo-label loss item also increases. 
\textcolor{black}{Therefore, the uncertain selection strategy caused the performance degradation and non-convergence problems of our model in the later stage of training.
On the contrary, the selection strategy of min-entropy has achieved the best effect.
In summary, in labels-at-client scenario, when the model's classification accuracy on the dataset is low, the low-entropy sample may be the wrong classification result. We should use the uncertain selection strategy to annotate unlabeled samples. Conversely, when the model has a high classification accuracy for the dataset, the low-entropy sample has a high probability of being the correct classification result. Therefore, we should use the min-entropy selection strategy.
}

As can be seen from Fig. \ref{ls-31} and Fig. \ref{ls-32}, for the CIFAR-10 and Fashion-MNIST datasets, we observe that the impact of the three selection strategies on performance is not obvious in labels-at-server scenario. 
\textcolor{black}{We guess that this may be because the global model fully absorbs the knowledge of the supervised model trained on the server-side labeled data.
Therefore, in this scenario, we randomly select a fixed-size pseudo-labeled dataset to reduce the computational overhead for training the unsupervised model.}


\begin{figure}[!t]
	\centering
	\large
	\subfigure[Labels-at-Client Scenario]{\includegraphics[width=0.47\linewidth]{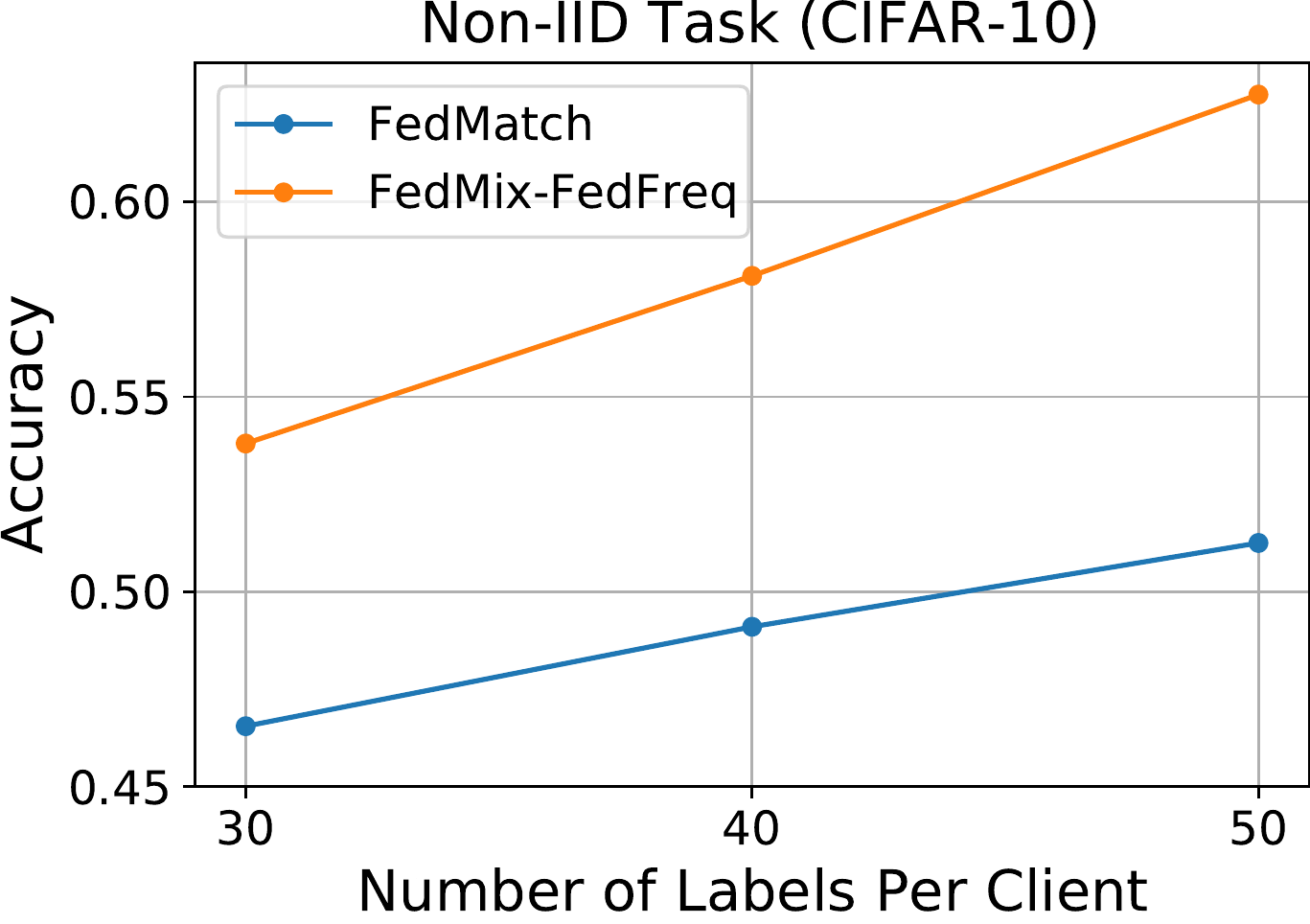}
		\label{lc-c-11}}
	\hfill
	\subfigure[Labels-at-Client Scenario]{\includegraphics[width=0.43\linewidth]{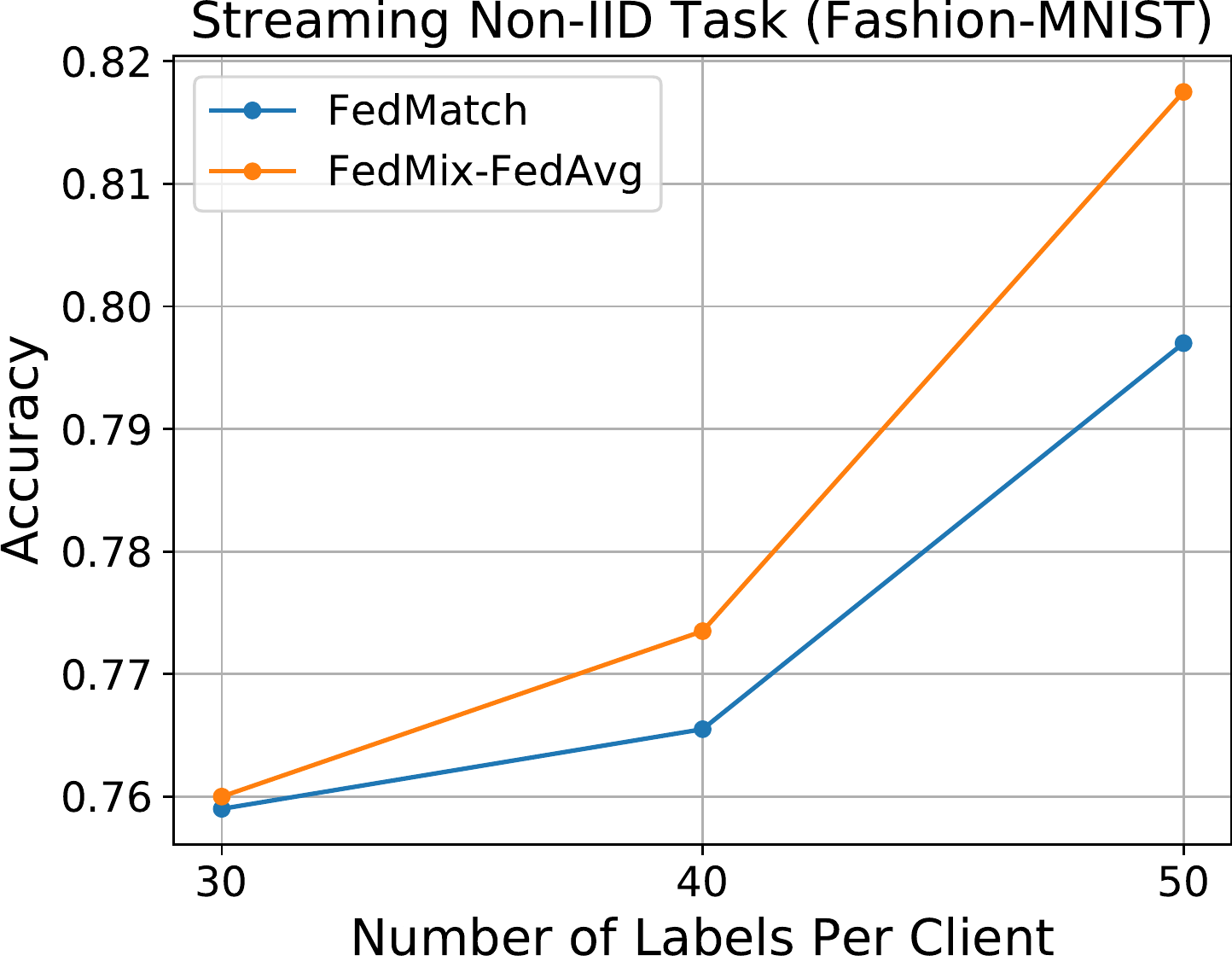}
		\label{lc-f-21}}

	\subfigure[Labels-at-Server Scenario]{\includegraphics[width=0.47\linewidth]{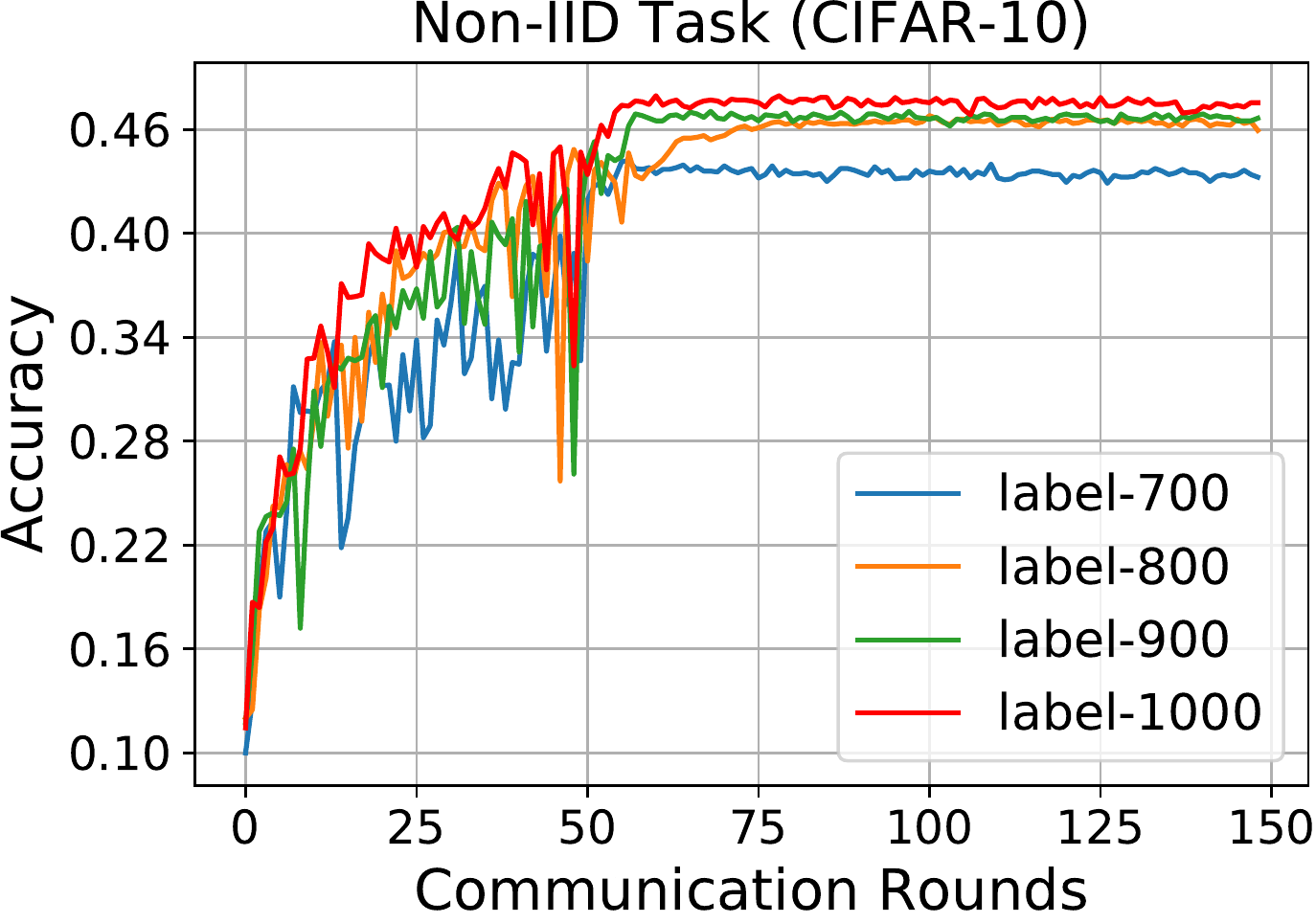}
		\label{ls-c-11}}
	\hfill
	\subfigure[Labels-at-Server Scenario]{\includegraphics[width=0.47\linewidth]{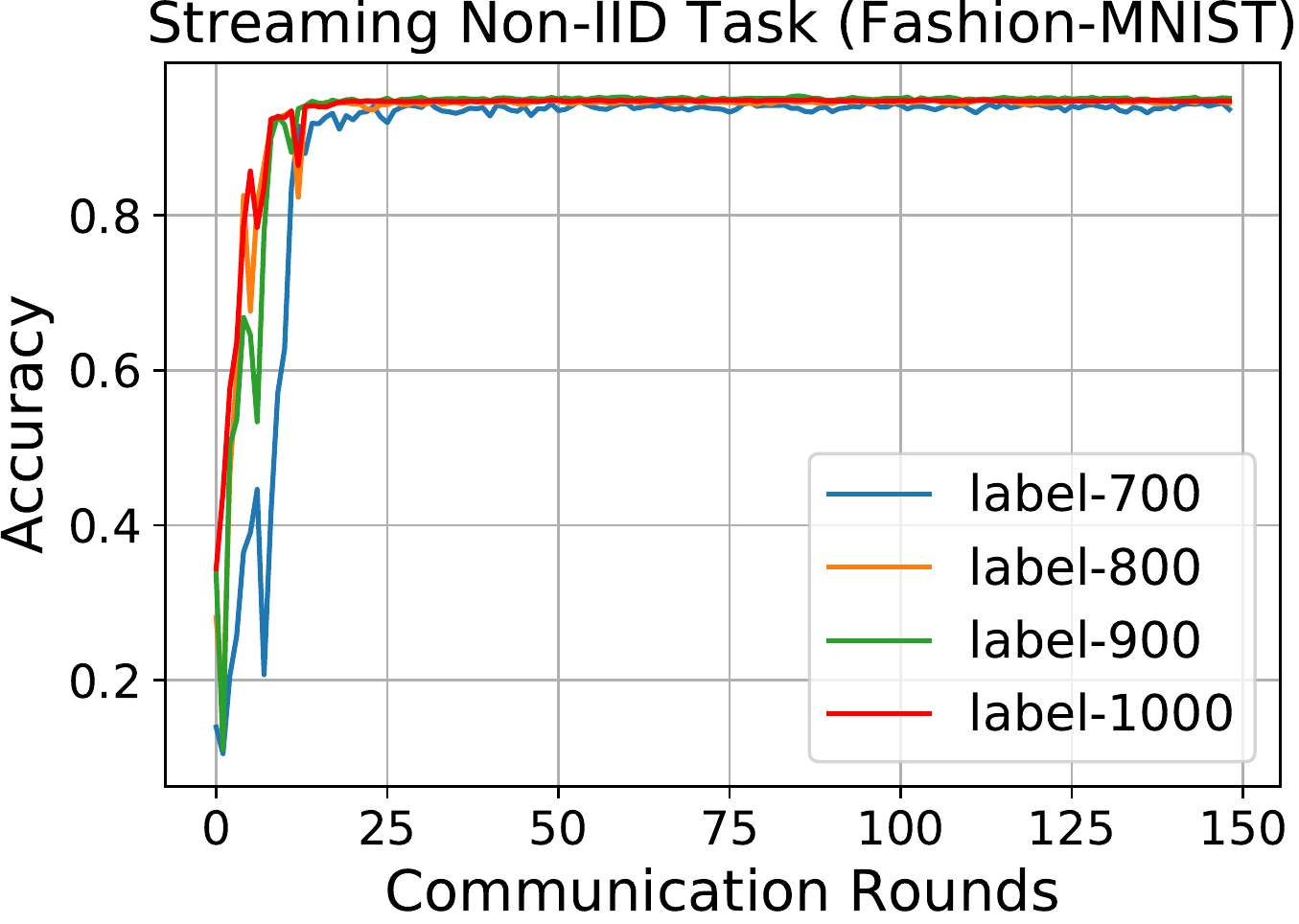}
		\label{ls-f-21}}
    \textcolor{black}{\caption{Performance comparison of different numbers of labeled samples for CIFAR-10 and Fashion-MNIST datasets in two scenarios.}}
	\label{F-2}
\end{figure}
\subsubsection{Performance comparison on different numbers of labeled samples in two scenario.}
\textcolor{black}{For the non-IID task of CIFAR-10 dataset in labels-at-client scenario, we explore the impact of the amount of labeled data held by each client on the model performance.} As shown in Fig. \ref{lc-c-11}, the performance of the global model largely depends on the amount of labeled data. The experimental result shows that $\mathrm{FedMix-FedFreq}$ is still better than $\mathrm{FedMatch}$ with a smaller number of labeled samples. 
For the Fashion-MNIST dataset, Fig. \ref{lc-f-21} shows streaming non-IID tasks with 50, 40, and 30 labeled samples per client. We observed that when the number of labeled decreased to 40, the model's performance decreased significantly, but it is also slightly higher than the baseline.

In the labels-at-server scenario, Fig. \ref{ls-c-11} shows the performance comparison of the proposed method in the case of different numbers of labeled samples at the server. The converged accuracy of our approach is 47\% with 800 labeled samples, which is 2\% higher than $\mathrm{FedMatch}$. However, when the number of labeled samples is reduced to 700, the accuracy of our model decreases significantly. Therefore, under the given setting and situation, we regard $N_s=800$ as the best setting for our method.
\textcolor{black}{On the other hand, as can be seen from Fig. \ref{ls-f-21}, reducing the number of labeled samples on the server side for the Fashion-MNIST dataset has no significant impact on accuracy.} 
\textcolor{black}{The reasons are: (1) Clothing images are easier to distinguish than physical images, and a small number of label samples can achieve higher performance.
(2) the model has learned more about unlabeled samples under the setting of local streaming data.}
These results demonstrate the effectiveness of our unsupervised training method.

\subsubsection{Performance comparison of different non-IID level in two scenario.}
\begin{figure}[!t]
	\centering
	\large
	\subfigure [Labels-at-Client Scenario]{\includegraphics[width=0.45\linewidth]{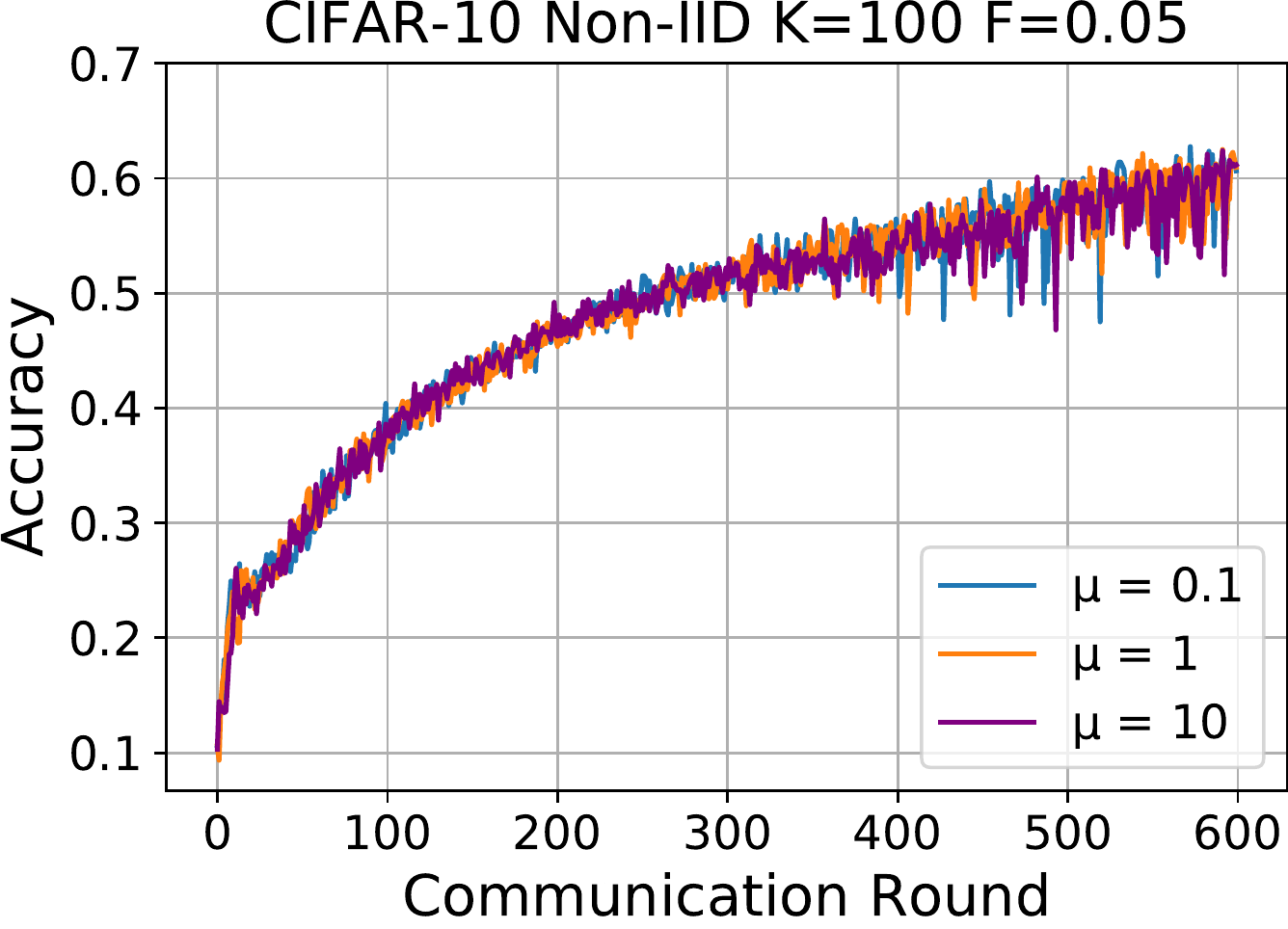}
		\label{lc-mu}}
	\hfill
	\subfigure[Labels-at-Server Scenario]{	\includegraphics[width=0.45\linewidth]{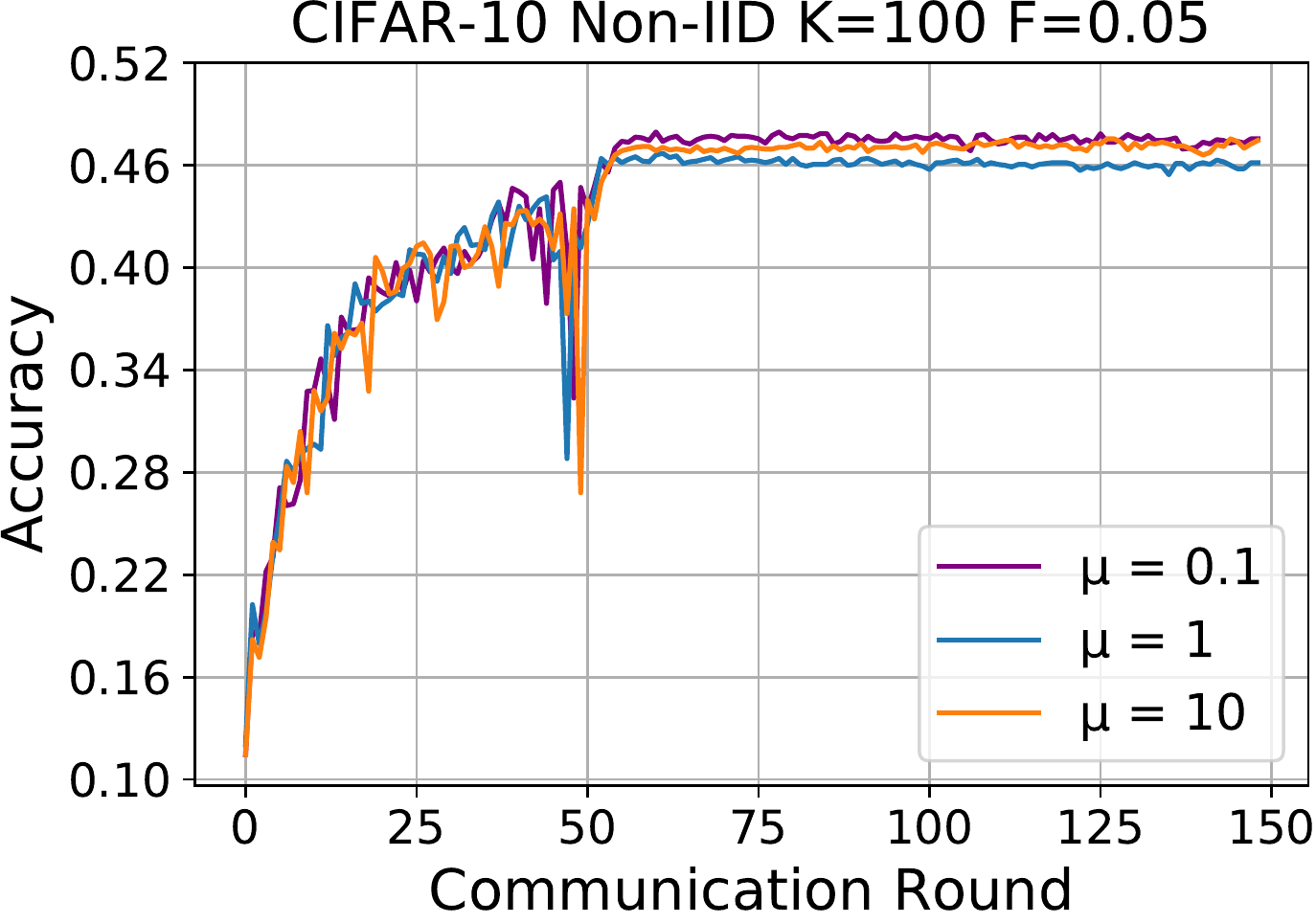}
		\label{ls-mu}}
	\caption{Performance comparison of different non-IID levels for the CIFAR-10 dataset in two scenarios.}
	\label{mu}
\end{figure}

Fig. \ref{mu} shows the performance comparison of the proposed method on different non-IID levels client data in two scenarios.
In our experiment, we let $\mu=0.1$ denote the highest non-IID level of the client data. In this case, as the value of $\mu$ increases, the local client data distribution becomes closer to the IID setting.
It can be seen from Fig. \ref{mu} that for different non-IID levels, our method can achieve stable accuracy. Meanwhile, the model convergence accuracy under $\mu=0.1,1,10$ settings does not differ by more than 1\%. Therefore, our method is not sensitive to the different levels of client data distribution, i.e., it is robust to different types of data distribution settings.

\subsubsection{Performance evaluation of Fashion-MNIST dataset streaming non-IID setting in two scenarios} To make a fair comparison with the $\mathrm{FedMatch}$, we let all clients participate in training each round, i.e., $F = 1.0$. Therefore, $\mathrm{FedFreq}$ aggregation rule cannot alleviate the non-IID in this case.
The reason is that under the setting of $F = 1.0$, the $\mathrm{FedFreq}$ aggregation algorithm becomes an average aggregation, i.e., the aggregation weight of each local model is the same. Thus, we use the mainstream $\mathrm{FedAvg}$ aggregation algorithm to verify the performance of our system. 

Table \ref{tab-3} shows the average model performance of 10 synchronized clients on streaming non-IID tasks. In the labels-at-client scenario, there is no significant performance difference between the $\mathrm{SL-FedAvg}$ model and the SSFL model. This means that the standard federated learning method does not fully use the knowledge of labeled data under the setting of streaming data. We speculate that under the setting of streaming data, the model may not train on new data adequately.
On the other hand, the labels-at-server scenario obtains higher accuracy than the labels-at-client scenario. 
The reasons for this phenomenon are: (1) the practical separation of supervised and unsupervised learning tasks dramatically improves the overall performance of our method. (2) the model makes full use of the knowledge of labeled data on the server.
Meanwhile, in both scenarios, our proposed method is superior to $\mathrm{SSL-FedAvg}$, and $\mathrm{FedMatch}$ and is close to federated learning under fully labeled data.

\begin{table}[!t]
	\centering
	\caption{Performance comparison of Fashion-MNIST dataset in two scenarios.}
	\begin{tabular}{|c|c|c|}\hline
		\multicolumn{3}{|c|}{Fashion-MNIST with 10 clients (K=10, F=1.0, A=3)}\\\hline
		\multicolumn{3}{|c|}{Accuarcy(\%)}\\\hline
		Methods&Labels-at-Client&Labels-at-Server\\\hline
		SL-FedAvg&83.02$\pm$0.44&N/A\\
		SSL-FedAvg&72.31$\pm$0.25&83.84$\pm$0.35\\
		FedMatch&78.45$\pm$0.31&92.48$\pm$0.33\\
		\textbf{FedMix-FedAvg}&\textbf{81.45$\pm$0.22}&\textbf{94.02$\pm$0.24}\\\hline
	\end{tabular}
	\label{tab-3}
\end{table}

\subsection{Discussion}
In this section, we further discuss the advantages and limitations of $\mathrm{FedMix}$ predicting image data in two SSFL scenarios. In the above, we conduct a comprehensive experiment to verify the effectiveness of our proposed method. Based on the above empirical results, the following observation results can be drawn.

\textbf{1) The performance of FedMix trained model under CIFAR-10 and Fashion-MNIST datasets is better than the mainstream SSFL baselines.}
Training a high-quality global model on a dataset with a small number of labeled samples is challenging in FL. The $\mathrm{FedMix}$ system utilizes a large number of unlabeled samples to reduce the model's dependence on labeled samples, and effectively obtains a high-precision prediction model. Specifically, we study the implicit contribution of the global model between iterative updates. On the other hand, based on information entropy, we propose two active learning strategies to screen high-quality pseudo-labels to improve the performance of the model \cite{settles1995active}.

\textbf{2) FedMix is robust to different levels of non-IID data.}
Statistical heterogeneity is another challenge faced by FL. We introduce the Dirichlet distribution function to simulate the client's non-IID data. Through experimental verification, our system is robust to different levels of non-IID data.

\textbf{3) Limitations.} By validating experimentally on two scenarios and two datasets of SSFL, and we observe that the active learning selection strategy shows different phenomena under different settings. For the labels-at-client scenario and CIFAR-10 dataset, the uncertainty strategy has a beneficial effect on the training model. However, for the Fashion-MNIST dataset, the uncertainty strategy caused model performance degradation and non-convergence. 
\textcolor{black}{This may be because the clothing images of the Fashion MNIST dataset are easier to distinguish than the physical images of CIFAR-10, and the performance of the model has a larger difference, resulting in different effects of different selection strategies.}
Moreover, in the labels-at-server scenario, the impact of the three selection strategies on model performance is not obvious.
\textcolor{black}{We guess that the reason for this phenomenon is that the supervised model of server-side training plays a leading role and ignores the unsupervised model of client-side training with different selection strategies.}
Therefore, the $\mathrm{FedMix}$ system shows the limitation of weak generalization ability.

\section{conclusion}\label{sec-8}
This paper presented a robust semi-supervised federated learning framework for aerial computing, accurately completing UAV image recognition tasks without revealing user privacy. In particular, we explored data availability (i.e., lack of data labels) and data heterogeneity (i.e., non-IID) in two realistic scenarios (i.e., labels-at-client and label-at-server). Specifically, to address the challenges caused by the lack of labeled data, we proposed the $\mathrm{FedMix}$ algorithm to achieve high-precision federated semi-supervised learning. To tackle the non-IID problem in FL, we proposed a novel aggregation algorithm, namely $\mathrm{FedFreq}$, which is based on client training frequency to aggregate. Simulation results show that our robust SSFL system is significantly better than existing solutions in performance under different settings. In future work, we will further improve the algorithm to maximize the use of unlabeled data. Furthermore, we will continue to strengthen the theory of SSFL so that it can be better applied in real-world scenarios.
\bibliographystyle{IEEEtran}
\bibliography{ref}

\end{document}